\documentclass[10pt,twocolumn,letterpaper]{article}

\usepackage[pagenumbers]{cvpr} % To force page numbers, e.g. for an arXiv version

\usepackage{graphicx}
\usepackage{amsmath}
\usepackage{amssymb}
\usepackage{booktabs}
\usepackage{multirow}
\usepackage{amsfonts}
\usepackage{siunitx}
\usepackage{svg}
\usepackage{caption}
\usepackage{float}

\usepackage[pagebackref,breaklinks,colorlinks]{hyperref}

\usepackage[capitalize]{cleveref}
\crefname{section}{Sec.}{Secs.}
\Crefname{section}{Section}{Sections}
\Crefname{table}{Table}{Tables}
\crefname{table}{Tab.}{Tabs.}

\def\cvprPaperID{*****} % *** Enter the CVPR Paper ID here
\def\confName{CVPR}
\def\confYear{2022}
\newcommand{\rev}[1]{\textcolor{black}{#1}}

\begin{document}

\title{FRoundation: Are Foundation Models Ready for Face Recognition?}

\author{Tahar Chettaoui\textsuperscript{1} \hspace{5mm} Naser Damer\textsuperscript{1,2} \hspace{5mm} Fadi Boutros\textsuperscript{1}
\vspace{0.35cm}\\
\textsuperscript{1}Fraunhofer IGD \hspace{3mm} \textsuperscript{2}Technische Universität Darmstadt
\vspace{0.1cm}\\
{\tt\small \{tahar.chettaoui, naser.damer, fadi.boutros\}@igd.fraunhofer.de}
}
\maketitle

\begin{abstract}
Foundation models are predominantly trained in an unsupervised or self-supervised manner on highly diverse and large-scale datasets, making them broadly applicable to various downstream tasks. In this work, we investigate for the first time whether such models are suitable for the specific domain of face recognition (FR). We further propose and demonstrate the adaptation of these models for FR across different levels of data availability, including synthetic data. Extensive experiments are conducted on multiple foundation models and datasets of varying scales for training and fine-tuning, with evaluation on a wide range of benchmarks. Our results indicate that, despite their versatility, pre-trained foundation models \rev{tend to} underperform in FR in comparison with similar architectures trained specifically for this task. However, fine-tuning foundation models yields promising results, often surpassing models trained from scratch, particularly when training data is limited. \rev{For example, after fine-tuning only on 1K identities, DINOv2 ViT-S achieved average verification accuracy on LFW, CALFW, CPLFW, CFP-FP, and AgeDB30 benchmarks of 87.10\%, compared to 64.70\% achieved by the same model and without fine-tuning.  While training the same model architecture, ViT-S, from scratch on 1k identities reached 69.96\%. With access to larger-scale FR training datasets, these performances reach 96.03\% and 95.59\% for the DINOv2 and CLIP ViT-L models, respectively. In comparison to the ViT-based architectures trained from scratch for FR, fine-tuned same architectures of foundation models achieve similar performance while requiring lower training computational costs and not relying on the assumption of extensive data availability. We further demonstrated the use of synthetic face data, showing improved performances over both pre-trained foundation and ViT models. Additionally, we examine demographic biases, noting slightly higher biases in certain settings when using foundation models compared to models trained from scratch. We release our code and pre-trained models' weights at \href{https://github.com/TaharChettaoui/FRoundation}{github.com/TaharChettaoui/FRoundation}}.
\end{abstract}

%%%%%%%%%%%%%%%%%%%%%%%%%%%%%%%%%%%%%%%%%%%%%%%%%%%%%%%%%%%%%%%%%%%%%%%%%%%%%%%%
\section{Introduction} \label{sec:intro}
Significant progress has been made in developing foundation models trained on extensive and diverse datasets recently \cite{oquab2024DINOv2learningrobustvisual, bao2022beitbertpretrainingimage, he2021maskedautoencodersscalablevision}. Once these models are trained, they can be adapted to a wide array of downstream tasks, providing a versatile basis. Inspired by the success of large language models (LLM) \cite{devlin2019bertpretrainingdeepbidirectional, brown2020languagemodelsfewshotlearners, touvron2023llamaopenefficientfoundation, chowdhery2022palmscalinglanguagemodeling, geminiteam2024geminifamilyhighlycapable}, similar large-scale foundation models have been explored for various perception tasks \cite{radford2021learningtransferablevisualmodels, oquab2024DINOv2learningrobustvisual, bao2022beitbertpretrainingimage, he2021maskedautoencodersscalablevision}. Often based on the ViT architecture \cite{dosovitskiy2021imageworth16x16words}, which has been shown to match or exceed the performance of traditional methods in large-scale image classification tasks, visual foundation models are becoming increasingly popular due to their strong generalization capabilities when trained on larger datasets. \rev{These models can be optimized to perform template extraction, using zero or few-shot learning approaches, making them highly versatile for biometric applications, where collecting a large set of training data, e.g. face images, is technically and legally challenging.} Although foundation models hold considerable promise for a wide range of applications, their adaptation for face recognition (FR) has not been explored in previous research, which \rev{motivates} this study. \rev{ In this work, we utilize visual foundation models to enhance performance for the downstream task of FR while minimizing reliance on extensive amounts of data, leveraging their pre-training on diverse datasets.}

To explore the potential of foundation models in FR, we evaluate the performance of different versions of two widely used foundation models, namely DINOv2 \cite{oquab2024DINOv2learningrobustvisual} and CLIP \cite{radford2021learningtransferablevisualmodels}. Although foundation models demonstrate strong generalization capabilities, our experiments show that they perform poorly compared to state-of-the-art (SOTA) FR models \cite{Deng_2022,wang2018cosfacelargemargincosine,ElasticFace,TransFace} on various benchmarks. To enhance their effectiveness for the downstream task of FR, we propose to fine-tune the considered foundation models using low-rank adaptation (LoRA) \cite{hu2021loralowrankadaptationlarge}. LoRA integrates trainable low-rank decomposition matrices into each transformer block while keeping the pre-trained model weights frozen. After fine-tuning on dedicated datasets, the LoRA layers adapt to the downstream task of FR, resulting in a significant increase in performance. For example, the smallest released \rev{pre-}trained models of DINOv2 and CLIP achieve average verification accuracies of 64.70\% and 82.64\%, respectively. After fine-tuning on CASIA-WebFace \cite{DBLP:journals/corr/YiLLL14a}, their accuracies increase to 90.94\% and 92.13\%, respectively. 

In addition to examining the performance of fine-tuned foundation models for FR, we compare them to a Vision Transformer (ViT) trained from scratch on different subsets of a \rev{small-scale training dataset, namely CASIA-WebFace.} \rev{The goal of this experiment is to leverage the versatility of foundation models, which are trained on extensive and diverse datasets, making them strong candidates for small-scale fine-tuning, as they benefit from the data on which they were previously trained. Additionally, we aim to address the challenges in data collection for the downstream task of FR, which can be tedious due to legal privacy regulations requiring strict consent, ethical considerations regarding individual rights, and technical limitations in collecting large and diverse training datasets.} \rev{Our experimental results demonstrate that, under conditions of low data availability, fine-tuned foundation models significantly outperform models trained from scratch, leveraging their pre-trained knowledge. On the other hand, as data availability increases, the performance of models trained from scratch becomes competitive.} \rev{To provide insight into foundation models performances when large-scale finetuning datasets are available and to provide comparison results to recent SOTA FR models, we finetuned/trained different instances of CLIP, DINOv2, and ViT on two large-scale datasets, MS1MV2 and WebFace4M, that are widely used in the literature.} \rev{Our results validated that models trained from scratch can eventually compete with or outperform fine-tuned foundation models when using large-scale datasets, highlighting the critical importance of selecting the appropriate training strategy based on dataset size.} 

\rev{With the growing use of synthetic data to develop FR models \cite{DBLP:journals/ivc/BoutrosSFD23, DBLP:journals/inffus/MelziTVKRLDMFOZZYZWLTKZDBVGFFMUG24, DBLP:conf/fgr/Otroshi-Shahreza24}, which enables privacy-preserving training, we explore the performance of these solutions and demonstrate that foundation models outperform models trained from scratch when both are trained on the same synthetic data.} \rev{The presented models in this paper are evaluated under cross-validation settings on mainstream challenging benchmarks, including ones with cross-age, cross-pose, and large-scale verification benchmarks. These evaluations are also enriched with demographic bias evaluation on racial face in the wild (RFW) by reporting verification accuracies as well as bias assessing metrics, e.g. skewed error ratio (SER) and standard deviations (STD). Our results on RFW, which are aligned with previous works \cite{PAMI_BIAS, DBLP:journals/corr/abs-1911-10692, DBLP:journals/inffus/MelziTVKRLDMFOZZYZWLTKZDBVGFFMUG24, DBLP:conf/cvpr/Gong0021, RFW}, report demographic bias in deep learning-based FR models.}

%%%%%%%%%%%%%%%%%%%%%%%%%%%%%%%%%%%%%%%%%%%%%%%%%%%%%%%%%%%%%%%%%%%%%%%%%%%%%%%%
\section{Related Work}
\rev{
Recent advances in FR performance have been primarily driven by advancing development in neural network architectures \cite{He2015DeepRL,dosovitskiy2021imageworth16x16words}, innovations in training loss functions \cite{wang2018cosfacelargemargincosine}, and the availability of large-scale training datasets \cite{guo2016ms, zhu2021webface260mbenchmarkunveilingpower, DBLP:conf/fgr/CaoSXPZ18}. 
The majority of recent FR studies \cite{wang2018cosfacelargemargincosine, Deng_2022, meng2021magfaceuniversalrepresentationface, ElasticFace, huang2020curricularfaceadaptivecurriculumlearning, kim2023adafacequalityadaptivemargin} employ ResNet-like \cite{He2015DeepRL} architectures, with some recent works \cite{DBLP:conf/bmvc/SunT22, DBLP:conf/cvpr/KimS0JL24} exploring ViT-based \cite{dosovitskiy2021imageworth16x16words} architectures. Most of these studies \cite{wang2018cosfacelargemargincosine, Deng_2022, meng2021magfaceuniversalrepresentationface, ElasticFace, huang2020curricularfaceadaptivecurriculumlearning, kim2023adafacequalityadaptivemargin} focus on innovations in training loss functions. FR training losses can be broadly categorized into two groups: metric learning (e.g., Triplet loss \cite{DBLP:conf/cvpr/SchroffKP15}) and multi-class classification learning (e.g., Softmax loss \cite{DBLP:conf/icml/LiuWYY16}). Metric learning losses \cite{DBLP:conf/nips/Sohn16,DBLP:conf/cvpr/SchroffKP15}, such as Triplet loss, explicitly encourage the model to learn discriminative feature representations by minimizing distances (e.g., Euclidean distance) between samples of the same label while maximizing distances between samples of different labels.
In contrast, margin-penalty softmax loss employs cross-entropy over a softmax layer to implicitly guide the model in learning discriminative features. This is achieved by deploying a margin penalty on the angle or cosine angle between samples and their respective class centers, aiming at pushing samples closer to their respective class centers and further away from other class centers. Innovations in margin-penalty softmax loss have focused on the geometric deployment of penalty margins, whether fixed \cite{wang2018cosfacelargemargincosine, Deng_2022} or adaptive \cite{meng2021magfaceuniversalrepresentationface, ElasticFace, huang2020curricularfaceadaptivecurriculumlearning, kim2023adafacequalityadaptivemargin}, yielding significant recognition improvements on mainstream benchmarks.
These advancements in FR research have been made possible by large-scale training datasets \cite{guo2016ms, zhu2021webface260mbenchmarkunveilingpower, DBLP:conf/fgr/CaoSXPZ18}, which enable the training of deep networks with millions of parameters. Notable datasets include CASIA-WebFace \cite{DBLP:journals/corr/YiLLL14a}, MS-Celeb-1M \cite{guo2016ms} and its subsets (MS1MV2 \cite{Deng_2022} and MS1MV3 \cite{DBLP:conf/iccvw/DengGZDLS19}), VGGFace2 \cite{DBLP:conf/fgr/CaoSXPZ18}, and WebFace260M \cite{zhu2021webface260mbenchmarkunveilingpower} with its subsets (WebFace42M, WebFace12M, and WebFace4M). However, these datasets are collected from the internet, raising discussion about the ethical use of such data in FR development \cite{DBLP:journals/ivc/BoutrosSFD23}. Such concerns, combined with the technical challenges of assembling large-scale datasets, have motivated researchers \cite{li2024id3identitypreservingyetdiversifieddiffusionmodels, kim2023dcfacesyntheticfacegeneration, DBLP:journals/tbbis/BoutrosHLSD24, boutros2023idifffacesyntheticbasedfacerecognition} to explore the use of synthetically generated data in FR development.
These challenges in collecting large-scale face datasets are among the key motivations for this work. As detailed later in this paper, we studied the impact of fine-tuning dataset size on the foundation model performances, providing insights into the model performance when extremely small subsets are available (e.g., 82k images of 1k identities) and in the case where large-scale datasets (e.g., 5.8M images of 85k identities) are accessible.
We also opted, as detailed later in this paper, to fine-tune foundation models on synthetically generated data, exploring the potential of such data in adapting foundation models for the FR task.}

\paragraph{Foundation models:}
Foundation models are defined by a substantial number of trainable parameters and are pre-trained on a large and diverse dataset, enabling them to adapt to a wide range of downstream tasks across various domains \cite{bommasani2022opportunitiesrisksfoundationmodels}. Vision foundation models are commonly structured around the use of Vision Transformers (ViTs) \cite{dosovitskiy2021imageworth16x16words} and rely on self-supervised learning (SSL) \cite{gui2024surveyselfsupervisedlearningalgorithms}. SSL is a technique that trains models to learn representations from unlabeled data and is essential for training ViTs, which tend to perform poorly on small datasets \cite{zhu2023understandingViTtrainsbadly}. Several high-performing vision foundation models specialized in various tasks have been released in recent years. Building on the success of the Segment Anything Model (SAM) \cite{kirillov2023segment}, which introduced a foundation model for object segmentation, SAM2 \cite{ravi2024sam2segmentimages} presents a model for real-time, promptable object segmentation in images and videos, achieving \rev{SOTA} performance. CLIP \cite{radford2021learningtransferablevisualmodels} learns visual concepts from natural language supervision and can be applied to any visual classification by providing the names of the visual categories to be recognized. DINOv2 \cite{oquab2024DINOv2learningrobustvisual} foundation models generate universal features suitable for both image-level visual tasks, such as image classification and video understanding, as well as pixel-level visual tasks, including depth estimation and semantic segmentation. 
% Also built on DINO, Grounding DINO \cite{} tackles open-set object detection, allowing it to identify arbitrary objects based on text queries. 

%% 2) model adaptation for downstream task, including different approaches for fine-tuning (zero-shot learning vs complete model fine-tuning vs adaptation).... you can here write about the application use cases where foundation model was adapted e.g. medical image, segmentation, classification 
Vision foundation models are characterized by their generalization ability due to massive training data, but they tend to show poor performance when applied to domain-specific settings \cite{wang2023sammeetsroboticsurgery}. To adapt them to a downstream task, multiple approaches were considered in the literature. An example of that is the AdaptFormer \cite{AdaptFormer2022} that replaces the MLP block in the transformer encoder with two identical MLP branches, where one mirrors the original network and the other introduces a lightweight module for task-specific fine-tuning, demonstrating significant improvements compared to fully fine-tuned models on downstream tasks. Another example is the ViT-Adapter \cite{chen2023visiontransformeradapterdense} that proposes a method to adapt plain ViTs for dense prediction tasks by injecting image-related inductive biases into the ViT and reconstructing fine-grained multi-scale features, yielding SOTA results on COCO test-dev. Another approach is to insert trainable rank decomposition matrices, called Low-rank adaptation (LoRA) \cite{hu2021loralowrankadaptationlarge} layers while freezing the pre-trained model weights. In this work, we choose LoRA as our foundation model adapter, as recent studies \cite{zhang2024learningadaptfoundationmodel, cui2024surgicaldinoadapterlearningfoundation, Chen_2023_ICCV, zanella2024lowrankfewshotadaptationvisionlanguage, wang2023sammeetsroboticsurgery} highlight the significant potential of LoRA for this purpose. For instance, integrating LoRA layers into DINOv2 has been successfully applied in the medical domain for two distinct tasks: capsule endoscopy diagnosis \cite{zhang2024learningadaptfoundationmodel} and depth estimation in endoscopic surgery \cite{cui2024surgicaldinoadapterlearningfoundation}, both of which have proven superior performance in their respective fields. In another work on adapting foundation models for multiple plant phenotyping \cite{Chen_2023_ICCV}, Chen et al. demonstrated that LoRA consistently outperforms decoder tuning in leaf counting and disease classification, with their method achieving high performance in both tasks, approaching the results of SOTA models. Zanella et al. \cite{zanella2024lowrankfewshotadaptationvisionlanguage} explored few-shot adaptation of Vision Language Models (VLMs) using LoRA, showing that their CLIP-based method not only achieves substantial improvements but also reduces training times. %Furthermore, \cite{wang2023sammeetsroboticsurgery} combined SAM and LoRA in the field of robotic surgery and demonstrated the capability of mask prediction without any prompts after fine-tuning.

%% 3) Foundation model in biometrics, and if there is any for face recognition (Arc2face but not really for representation learning)
The application of foundation models in biometrics is still very limited, with only a few recent works starting to investigate their potential. For example, Iris-SAM \cite{farmanifard2024irissamirissegmentationusing}, which is based on the foundation model SAM \cite{kirillov2023segment}, fine-tunes it on ocular images for iris segmentation, achieving an average segmentation accuracy that surpasses the best baseline by a substantial margin of 10\% on the ND-IRIS-0405 dataset. Arc2Face \cite{papantoniou2024arc2facefoundationmodelidconsistent} is an identity-conditioned face foundation model that generates photo-realistic images based on the ArcFace \cite{Deng_2022} embedding of a person. To showcase the performance of the generated data, they train a FR model on synthetic images from their model, achieving superior performance compared to existing synthetic datasets \cite{DBLP:journals/ivc/BoutrosSFD23}. Recognizing the immense potential of foundation models across diverse tasks, this study uncovers new perspectives by exploring their adaptation for FR, a path \rev{that}, to the best of our knowledge, \rev{has been} unexplored until now.

%%%%%%%%%%%%%%%%%%%%%%%%%%%%%%%%%%%%%%%%%%%%%%%%%%%%%%%%%%%%%%%%%%%%%%%%%%%%%%%%
\section{Methodology}
This section presents our approach FRoundation, for optimizing vision foundation models for FR. This section starts by providing details on the selected baseline foundation models, CLIP \cite{radford2021learningtransferablevisualmodels} and DINOv2 \cite{oquab2024DINOv2learningrobustvisual}. Then, we provide details on the adaptation mechanism used to optimize foundation models for downstream tasks. Finally, we conclude by describing the extension of pre-trained foundation models with LoRA for FR.

\begin{figure*}[ht!]
  \centering
   \includegraphics[width=1\linewidth]{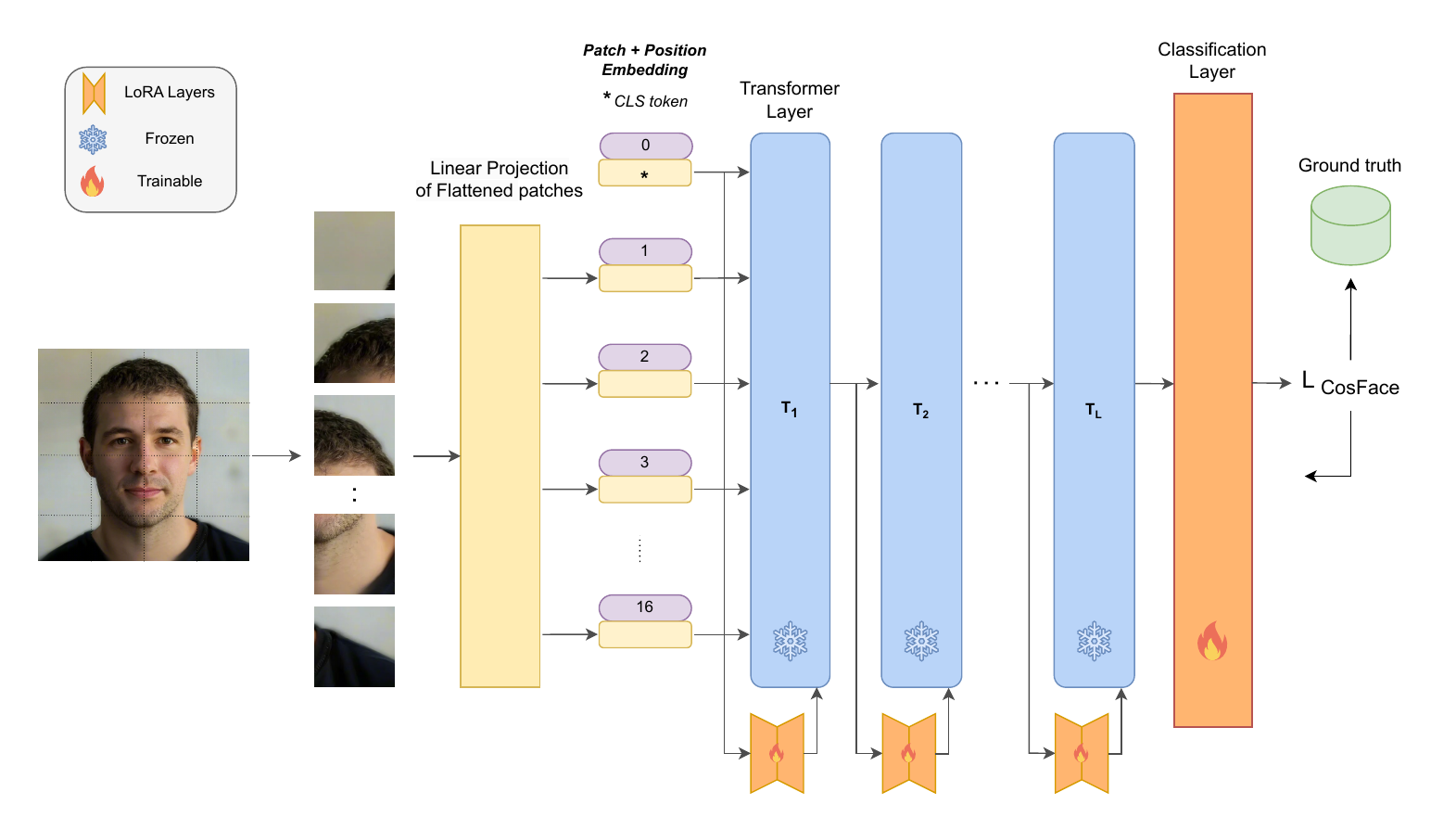}
   \caption{ViT Training Pipeline with LoRA Adapters. \rev{The facial image is divided into patches, which are mapped to patch embeddings via a linear projection. A class token is added, and positional embeddings are incorporated to maintain spatial information. This sequence of embedding vectors is then input into the encoder.} The transformer layers remain fixed during training, while trainable LoRA layers are introduced to fine-tune the model. Each LoRA layer operates independently within the transformer layers and possesses its own set of weights.}
   \label{fig:ViTlora}
\end{figure*}

\subsection{Baseline Foundation Models} \label{sec:Baseline Foundation Models}
We selected two \rev{SOTA} foundation models, CLIP and DINOv2, to conduct our studies in this paper. 
These models proved to be generalizable across different downstream tasks \cite{oquab2024DINOv2learningrobustvisual, radford2021learningtransferablevisualmodels} and achieved very competitive results with zero-shot learning. 
% For example, CLIP and DINOv2 achieved xx\% and yy\% accuracies on ImageNet, respectively which is very competitive to zz\% achieved by the SOTA approach \cite{}. 
Previous works \cite{zhang2024learningadaptfoundationmodel, cui2024surgicaldinoadapterlearningfoundation} also showed that fine-tuning these models using, for example, LoRA, could lead to SOTA performance on specific downstream tasks.

%%%%%%%%%%%%%%%%%%%%%%%%%%%%%%%%%%%%%%%%%%%%%%%%%%%%%%%%%%%%%%%%%%%%%%%%%%%%%%%%
\subsubsection{DINOv2} 
% DINOv2 was built on DINO \cite{} and iBOT \cite{} and proposed several acceleration techniques to improve the performance of a small model by distilling the knowledge from a large model.
%Introduced by Oquab et al. \cite{oquab2024DINOv2learningrobustvisual}, DINOv2 comprises a series of self-supervised image encoders pre-trained on a large curated dataset, namely the LVD-142M dataset. The dataset was created as part of this initiative, using an automated pipeline to assemble a dedicated, diverse, and curated collection of images. 

DINOv2 \cite{oquab2024DINOv2learningrobustvisual} is a self-supervised image encoder trained on a large curated dataset, namely the LVD-142M dataset. The dataset was created as part of this initiative, using an automated pipeline to assemble a dedicated, diverse, and curated collection of images. The model network architecture follows a student-teacher framework based on vision transformers (ViT) \cite{dosovitskiy2021imageworth16x16words} that learns features at the image and patch levels by combining DINO \cite{caron2021emergingpropertiesselfsupervisedvision} and iBOT \cite{zhou2022ibotimagebertpretraining} losses. The image-level objective, inspired by DINO, is derived from the cross-entropy loss between features extracted from the student and teacher networks. These features are taken from the ViT class token and are based on different crops of the same image. For the patch-level objective, random input patches are masked and sent to the student, while remaining visible to the teacher. The student’s iBOT head processes the masked tokens while the teacher’s iBOT head processes the corresponding visible tokens, leading to the calculation of the loss term. Additionally, several contributions were made to accelerate and stabilize large-scale training. As a result, a ViT model with 1B parameters was trained and distilled into a series of smaller models that surpass the best available general-purpose features on most of the benchmarks at image and pixel levels \cite{oquab2024DINOv2learningrobustvisual}, making it a top candidate as a foundation model for FR.

%%%%%%%%%%%%%%%%%%%%%%%%%%%%%%%%%%%%%%%%%%%%%%%%%%%%%%%%%%%%%%%%%%%%%%%%%%%%%%%%
\subsubsection{CLIP} 
Introduced by Radford et al. \cite{radford2021learningtransferablevisualmodels}, Contrastive Language-Image Pretraining (CLIP) is a multimodal foundation model that jointly learns from visual and textual modalities. It leverages a large dataset comprising images paired with text description, enabling the model to learn and relate visual information to textual context, and vice versa. The architecture consists of two separate encoders to process image and text inputs, respectively. During training, CLIP employs a contrastive learning approach, maximizing the cosine similarity between feature representations obtained from image-text pairs while minimizing it for negative samples. This allows the model to effectively capture the relationship between images and their corresponding textual descriptions. The training process involves a large-scale training dataset, facilitating the model's ability to generalize across a variety of visual and textual tasks. In this work, we will focus exclusively on employing the image encoder as we aim to obtain feature representation from face images for the face verification task.

%%%%%%%%%%%%%%%%%%%%%%%%%%%%%%%%%%%%%%%%%%%%%%%%%%%%%%%%%%%%%%%%%%%%%%%%%%%%%%%%
\subsection{Fine-tuning with LoRA} \label{sec:Fine-tuning with LoRA}
In this work, we utilize Low-rank adaptation (LoRA) \cite{hu2021loralowrankadaptationlarge} to fine-tune the considered foundation model. LoRA was initially developed to fine-tune LLMs for specific downstream tasks, but it can be applied to any neural network with dense layers. Its development was inspired by \cite{aghajanyan2020intrinsicdimensionalityexplainseffectiveness}, which demonstrates that a low-dimensional reparameterization can be as effective for fine-tuning as the full parameter space. This indicates that pre-trained models actually reside on a low intrinsic dimension. Building on this concept, LoRA freezes the pre-trained model weights and inserts trainable rank decomposition matrices into each layer of the \rev{pre}-trained transformer architecture. For a pre-trained weight matrix $W_0 \in \mathbb{R}^{d \times k}$, LoRA utilizes a low-rank decomposition to restrict its update by $W_0 + \Delta W = W_0 + BA$ where $B \in \mathbb{R}^{d \times r}$ and $A \in \mathbb{R}^{r \times k}$ with the rank $r \ll min(d,k)$. $W_0$ does not receive gradient updates during the training process, while only $A$ and $B$ are updated. When fine-tuning, this approach significantly reduces the number of trainable parameters compared to fine-tuning all the model parameters, while also not introducing any additional inference latency. The latter is achieved by computing $W = W_0 + BA$. After fine-tuning the model, the adapter weights $BA$ are merged with the base model weight $W_0$ to compute the final model weights $W$.
%This is beneficial because it allows for faster processing times and more efficient use of resources, enabling the model to operate effectively in real-time applications and on devices with limited memory capacity.

During the forward pass, the low-rank matrix product $BA$ is scaled by $\dfrac{\alpha}{r}$ where $\alpha$ is a constant. When optimizing with Adam \cite{kingma2017adammethodstochasticoptimization}, tuning $\alpha$ is roughly the same as tuning the learning rate \cite{hu2021loralowrankadaptationlarge}. This scaling factor causes gradient collapse as the rank increases, resulting in larger ranks performing no better than smaller ones \cite{kalajdzievski2023rankstabilizationscalingfactor}. To tackle this issue, the rank-stabilized LoRA (rsLoRA) \cite{kalajdzievski2023rankstabilizationscalingfactor} method proposes to scale the low-rank matrix with $\dfrac{\alpha}{\sqrt{r}}$. Gradients do not collapse with rsLoRA, and training with higher ranks has been experimentally validated to improve performance. This method allows for an effective compute/performance trade-off, where higher ranks can be used to achieve higher performance at the cost of increased training computation.

%%%%%%%%%%%%%%%%%%%%%%%%%%%%%%%%%%%%%%%%%%%%%%%%%%%%%%%%%%%%%%%%%%%%%%%%%%%%%%%%
\subsection{FRoundation}  \label{sec:FRoundation}

\begin{figure}[ht!]
  \centering
   \includegraphics[width=0.7\linewidth]{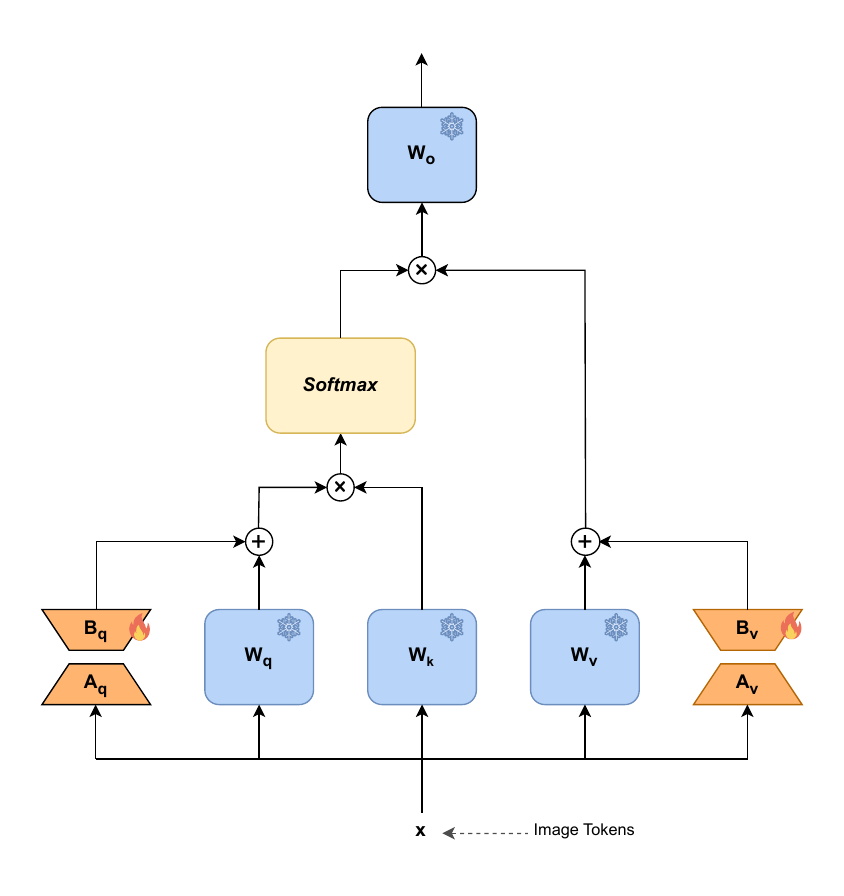}
   \caption{LoRA Integration in Multi-Headed Self-Attention Block. We implement LoRA to $q$ and $v$ projection layers in each transformer block.}
   \label{fig:mhalora}
\end{figure}

Utilizing \rev{pre}-trained foundation models for FR leads to suboptimal results, as will be presented in Section \ref{sec:eval_1}. Thus, we propose to fine-tune the considered foundation models using LoRA. This requires extending the \rev{pre}-trained ViT models with LoRA layers before fine-tuning them on dedicated datasets. It is possible to apply LoRA to the $q$, $k$, $v$, and $o$ projection layers, which refer respectively to the query, key, value, and output projection matrices in the self-attention module of ViT architecture \cite{dosovitskiy2021imageworth16x16words}. Following \cite{hu2021loralowrankadaptationlarge}, we adapt the weight matrices of $q$ and $v$ only, as it was shown that adapting their respective weights $W^{q}$ and $W^{v}$ yields the best results on different downstream tasks \cite{hu2021loralowrankadaptationlarge}.

The transformer encoder consists of alternating layers of Multiheaded Self-Attention (MSA) and multilayer perceptron (MLP) blocks. Layer Normalization (LN) is applied before every block and residual connections after every block. LoRA is applied exclusively to the attention weights, leaving the MLP unchanged for both simplicity and parameter efficiency \cite{hu2021loralowrankadaptationlarge}. As illustrated in Fig \ref{fig:ViTlora}, given a facial image, the image is divided into non-overlapping patches which are then mapped into patch embeddings using a linear projection layer. Additionally, a learnable embedding known as the class (CLS) token \cite{dosovitskiy2021imageworth16x16words} is appended to the sequence of embedded patches. The goal of the CLS token is to serve as the image representation, which we utilize to obtain feature representation from input face samples. Then, position embeddings are incorporated into the patch embeddings to preserve positional information. The resulting sequence of embedding vectors is used as input for the encoder. In MSA, we run $k$ heads in parallel, each with its own set of $q$, $k$, and $v$. In each head, LoRA layers operate separately and have their own distinct weights. As shown in Fig \ref{fig:mhalora}, for an embedding $x$, the computation of the $q$, $k$, and $v$ projection layers in head $i$ are:
\begin{equation} 
\begin{split}
  Q_i & = W^{q}_i x + B^{q}_i A^{q}_i x,  \\  
  K_i & = W^{k}_i x,  \\ 
  V_i & = W^{v}_i x + B^{v}_i A^{v}_i x \\
\end{split}
\label{eq1}
\end{equation}

$W^{q}_i$, $W^{k}_i$, and $W^{v}_i$ are frozen projection layers for $q$, $k$, and $v$, respectively, while $A^{q}_i$, $B^{q}_i$, $A^{v}_i$, and $B^{v}_i$ are the trainable LoRA layers. Then the attention scores are computed for all heads using a scaled dot-product attention mechanism. The attention output for head $i$ is:
% review math symbols 
\begin{equation} 
\begin{split}
  Attention(Q_i, K_i, V_i) & = Softmax \left( \dfrac{Q_i K_i^{T}}{\sqrt{d_k}} \right) V_i,  \\  
\end{split}
\label{eq2}
\end{equation}
Where the scaling factor $d_k$ represents the dimension of the key vectors.
The output of all heads is concatenated along the feature dimension and passed through the projection layer O:
% review math symbols 
\begin{equation} 
\begin{split}
MutliHead(Q, K, V) & = Concat(head_1, .., head_k) W^{o}.\\
\end{split}
\label{eq3}
\end{equation}

The output of the projection layer $O$ is then passed to the MLP, completing the execution of a single transformer block. $L$ transformer layers are used to transform the image tokens into feature representations $t^{l}$ where $l$ denotes the output of the $l$-th transformer block. Each $l$-th transformer layer processes the output vectors of the previous layer. The final hidden state of the CLS token is employed as feature representation. To optimize the considered foundation models for FR, we proposed to fine-tune the models with the widely adapted margin-penalty softmax loss for FR \cite{wang2018cosfacelargemargincosine, Deng_2022}. Specifically, we extended the architecture with an additional multi-class classification layer and utilized CosFace loss \cite{wang2018cosfacelargemargincosine} as a margin-penalty softmax loss.  

\begin{table*}[ht]
%\captionsetup{justification=centering}
\caption{The achieved verification performances by DINOv2 and CLIP on several evaluation benchmarks. The results are reported in (\%) on the small benchmarks and as average accuracies. The results of the first row are achieved by the ViT-S model trained from scratch on MS1MV2 and provided as a reference. The DINOv2 dataset blurred faces during training, as discussed in Section \ref{sec:eval_1}, which causes the gap in evaluation accuracies between the two models.
%Are Foundation Models ready for Face Recognition? The DINOv2 dataset blurred faces during training, as explained in Section \ref{sec:eval_1}, which causes the gap in evaluation accuracies between the two models.
}
\label{table:FM_nofine-tuning}
\centering
\resizebox{\linewidth}{!}{%
\begin{tabular}{cc|cccccc|cccccc}
\multirow{3}{*}{} &  &  &  &  &  & & &  &  & &  & \\
                  Method & Backbone & LFW & CFP-FP & AgeDB30 & CALFW & CPLFW & Avg. &\multicolumn{3}{c}{IJB-B} & \multicolumn{3}{c}{IJB-C} \\
                  &  &  &  &  &  & & & $10^{-3}$ & $10^{-4}$ & $10^{-5}$  & $10^{-3}$ & $10^{-4}$ & $10^{-5}$\\ \hline
                Baseline (MS1MV2) & ViT-S & 99.73 & 95.44 &  97.42 & 95.95 & 92.35 & 96.18 & 95.89 & 92.85 & 79.68 & 96.76 &  94.51 & 88.30\\ \hline \hline

\multirow{5}{*}{DINOv2} 
                & ViT-S &  78.73 & 71.15 & 54.70 & 59.47 & 59.48 & 64.70 & 14.66 & 5.90 & 2.33 & 18.10  & 7.44 & 2.87 \\
                & ViT-B &  80.22 & \bf{72.64} & 56.45 & 59.77 & 63.10  & 66.44 & 15.13 & 5.77 & 2.44  & 18.21 & 6.90 & 2.59 \\
                & ViT-L &  \bf{80.37} & 71.97 &  55.25 & \bf{60.52} & \bf{64.33}  & 66.49 & 17.18 & 6.44 & 2.52 & \bf{20.36}  & 7.84 & 2.77 \\
                & ViT-G &  80.32 & 71.97 & \bf{58.62} & 59.53 & 64.20  & \bf{66.93} & \bf{17.59} & \bf{6.76} & \bf{2.79} &  19.94 & \bf{8.05} & \bf{3.23} \\ \hline \hline
  
\multirow{4}{*}{CLIP} 
                & ViT-B/32  &  94.03 & 84.91 & 70.72 & 76.13 & 77.47  & 80.65 & 39.36 & 20.58 & 9.38 & 43.73 & 25.02  & 11.98 \\
                & ViT-B/16  &  93.33 & 88.86 & 74.67 & 77.13 & 79.23  & 82.64 & 49.19 & 27.79 & 11.93 & 52.21 & 32.40 & 16.34 \\
                & ViT-L/14  &  95.90 & 90.66 & \bf{79.82} & \bf{83.10} & \bf{82.73}  & \bf{86.44} &\bf{62.72} & \bf{40.90} & \bf{20.52} & \bf{64.74} & \bf{44.69} & \bf{25.23} \\
                & ViT-L/14@336  & \bf{96.30} & \bf{91.26} & 79.80 & 81.83 & 82.30 & 86.30 & 61.88 & 39.69 & 18.86 &  64.30 & 43.68  & 24.13 \\  

\end{tabular}}
\end{table*}

%%%%%%%%%%%%%%%%%%%%%%%%%%%%%%%%%%%%%%%%%%%%%%%%%%%%%%%%%%%%%%%%%%%%%%%%%%%%%%%%

% baseline small 
% 22,056,576 

% baseline_large
% 304,368,640

% dinov2_small
% before lora 22,056,576     
% after lora 22,351,488
% trainable 294,912 ==> 1.32

% dinov2_large
% before lora 304,368,640 
% after lora 1,572,864
% trainable 1,572,864  ==> 0.51 

% clip_base
% before lora 86,192,640 
% after 86,782,464
% trainable 589,824 ==> 0.68

% clip_large 
% before lora 303,966,208
% after lora 30,5539,072
% trainable 1,572,864 ==> 0.51

\section{Experimental setups}  \label{sec:experimental}
\paragraph{Evaluation Datasets} \label{sec:experimental_datasets} To explore the capabilities of foundation models for FR, we assess their performance using the following benchmarks: Labeled Faces in the Wild (LFW) \cite{huang:inria-00321923}, Celebrities in Frontal-Profile in the Wild (CFP-FP) \cite{c3517bca662f4193a58fd8f9e3145c8f}, AgeDB30 \cite{moschoglou2017agedb}, Cross-age LFW (CA-LFW) \cite{DBLP:journals/corr/abs-1708-08197}, CrossPose LFW (CP-LFW) \cite{CPLFWTech}. The results are reported on each benchmark as verification accuracies in (\%) and following their official evaluation protocol.
In addition, we evaluated on large-scale evaluation benchmarks, IARPA Janus Benchmark-B (IJB-B) \cite{inproceedingsijbb}, and IARPA Janus Benchmark–C (IJB-C) \cite{DBLP:conf/icb/MazeADKMO0NACG18}. For IJB-C and IJB-B, we used the official 1:1 mixed verification protocol and reported the verification performance as true acceptance rates (TAR) at false acceptance rates (FAR) of 1e-3, 1e-4, and 1e-5. All these benchmarks were chosen as they are widely used to benchmark the latest FR developments and provide a comprehensive variation of use-cases \cite{Deng_2022,wang2018cosfacelargemargincosine,ElasticFace,TransFace}. \rev{All results in the paper are reported using cross-validation settings, ensuring no overlap between the training and testing datasets. This approach aligns with SOTA works \cite{Deng_2022, ElasticFace, meng2021magfaceuniversalrepresentationface} in FR, by utilizing datasets such as CASIA-WebFace, MS1MV2, and WebFace4M  for training/finetuning while reporting verification results on mainstream evaluation benchmarks, including LFW, CFP-FP, AgeDB30, CALFW, CPLFW,  CFP-FP, IJB-C and IJB-B. }

\paragraph{Training and Fine-tuning Datasets} 
To fine-tune the considered foundation models, we employ the CASIA-WebFace \cite{DBLP:journals/corr/YiLLL14a} dataset, consisting of 0.5 million images and 10\rev{K} identities. We also conduct large-scale fine-tuning on the MS1MV2 \cite{Deng_2022} and WebFace4M \cite{zhu2021webface260mbenchmarkunveilingpower} datasets. MS1MV2 is a refined version of  MS-Celeb-1M \cite{guo2016ms} by \cite{Deng_2022}, containing 5.8M images of 85K identities.  WebFace4M is a subset of the WebFace260M dataset \cite{zhu2021webface260mbenchmarkunveilingpower}, consisting of 200\rev{K} identities and 4 million images. \rev{We also investigate the performance of the selected foundation models when trained using synthetic data, focusing on a latent diffusion models-based approach, IDiff-Face \cite{boutros2023idifffacesyntheticbasedfacerecognition}, and a generative adversarial network (GAN)-based approach, SFace2 \cite{DBLP:journals/tbbis/BoutrosHLSD24}. For both synthetic datasets, we conducted training on 10K identities, using 50 images per identity.} The images in the training and testing dataset are aligned and cropped to $112 \times 112$ as described in \cite{Deng_2022} using five landmark points extracted by the Multi-task Cascaded Convolutional Networks (MTCNN) \cite{MTCNN}.

\paragraph{Model architectures}
%: ViT-S has 384 dimensions and 6 heads, ViT-B has 768 dimensions and 12 heads, ViT-L has 1024 dimensions and 16 heads, while ViT-g has 1536 dimensions and 24 heads. 
DINOv2 \cite{oquab2024DINOv2learningrobustvisual} officially released four ViT architectures, small (Vi\rev{T}-S), base (Vi\rev{T}-B), large (Vi\rev{T}-L), and giant (Vi\rev{T}-G). ViT-S, Vi\rev{T}-B, Vi\rev{T}-L\rev{,} and Vi\rev{T}-G contain 2\rev{2}M, 86M, 0.3B\rev{,} and 1.1B parameters, respectively. All models use a patch size of 14 \rev{but differ in embedding dimensions and the number of attention heads, with ViT-S having an embedding dimension of 384 and 6 heads, ViT-B having an embedding dimension of 768 and 12 heads, ViT-L having an embedding dimension of 1024 and 16 heads, and ViT-g having an embedding dimension of 1536 and 24 heads.} On the other hand, CLIP \cite{radford2021learningtransferablevisualmodels} offers 4 different models with 2 architectures: base and large. The base model ViT-B of CLIP contains 86M parameters and has 2 variants with different patch sizes, 16 \rev{(ViT-B/16)} and 32 \rev{(ViT-B/32)}. The large model ViT-L/14 has 0.3 billion parameters and includes a variant\rev{, namely ViT-L/14@336,} that was pre-trained at a higher resolution of 336 pixels for one additional epoch to boost performance \cite{DBLP:journals/corr/abs-1906-06423}. \rev{The ViT-B model has a width of 768 and 12 attention heads, while the ViT-L model has a width of 1024 and 16 attention heads.} We will start by evaluating the performance of all models in Section \ref{sec:eval_1}. For further detailed experiments, we focus on the following models: ViT-S for DINOv2 and \rev{ViT-B/16} for CLIP, as these are the smallest models released by the corresponding authors, which makes our detailed experiments viable (given hardware and time limitations). We also investigate the performance of the larger models in Section \ref{sec:eval_2} and choose ViT-L for CLIP and DINOv2 for a fairer comparison, as the largest model, namely the giant model of DINOv2, has 1.1 billion parameters compared to 0.3 billion parameters for DINOv2 and CLIP ViT Large. 

\paragraph{Training Settings} 
We utilize the CosFace \cite{wang2018cosfacelargemargincosine} loss function to train all \rev{models presented} in this paper with a margin penalty of $0.3$ and scale factor of $64$, following \cite{wang2018cosfacelargemargincosine}, as well as other works analyzing different building blocks of FR \cite{DBLP:journals/tbbis/BoutrosHLSD24}.
During the fine-tuning, we used AdamW \cite{loshchilov2019decoupledweightdecayregularization} optimizer with a weight decay of $0.05$. We train for $40$ epochs (on \rev{CASIA-WebFace}) and for $30$ epochs (on MS1MV2 and WebFace4M), with a batch size set to $512$ \cite{Deng_2022}. The initial learning rate is set to $0.0001$ and is updated using a Cosine Learning Rate scheduler \cite{loshchilov2017sgdrstochasticgradientdescent}. Additionally, the LoRA rank is set to $16$ for all applicable experiments. The images are resized to $224 \times 224$ pixels to adapt to the image resolution initially used by DINOv2 and CLIP during training. For data augmentation, we apply horizontal flipping and RandAug \cite{Randaugment_CVPR} with $4$ operations and a magnitude of $16$, following \cite{DBLP:conf/icb/BoutrosKFKD23}. \rev{During training and following \cite{Deng_2022, ElasticFace}, we explore efficient face verification datasets (e.g. LFW, CALFW, CPLFW, CFP-FP, AgeDB) to track and check the convergence status of the model. This has been performed after each epoch. It is important to note that for the RFW \cite{RFW} evaluation, models trained on MS1MV2 are not subject to full cross-validation due to the identity overlap between RFW and MS1MV2, which is a cleaned version derived from MS-Celeb-1M \cite{guo2016ms}, as stated in \cite{RFW}.} We additionally trained $12$ instances of Vi\rev{T}, Vi\rev{T}-S, and Vi\rev{T}-L, from scratch on different subsets of CASIA-WebFace as well as on MS1MV2 and WebFace4M. These models are noted as the baseline.

\paragraph{Computational Cost}
\rev{We utilized two foundation models in this paper, each with different base architectures. For DINOv2, we utilized ViT-S and ViT-L, containing 22M and 0.3B parameters, respectively. For CLIP, we utilized ViT-B and ViT-L, containing 86M and 0.3B parameters, respectively. Given that these models utilize 32 floating-point (4-byte) values to represent each parameter, the required memory footprint for each model is 4 times the number of parameters.}
\rev{To evaluate computational time during inference, we measure the models' latency on a Quadro RTX 6000 GPU. Specifically, the ViT-S and ViT-L variants of DINOv2 require 6.993 ms and 13.614 ms, respectively, to process a single image on this hardware. When fine-tuning the models, we attach LoRA layers to the various considered model architectures. These layers introduce computational overhead, as the forward pass of the model becomes more complex due to the need to process the additional weights. As a result, the forward pass time for the DINOv2 ViT-S and ViT-L increases to 9.862 ms and 19.502 ms, respectively.} \rev{For the CLIP variants, namely ViT-B and ViT-L, they require 8.657 ms and 12.74 ms, respectively, which increase to 12.498 and 18.817 ms when LoRA layers are added.} 
\rev{While models fine-tuned with LoRA exhibit a slower forward pass during training, they benefit from the fact that only the additional LoRA layers are updated during the backpropagation process. For DINOv2, ViT-S (22M parameters) and ViT-L (0.3B parameters) incorporate an additional 0.3M and 1.5M parameters due to LoRA, respectively, which corresponds to only 1.32\% and 0.51\% of the total trainable parameters. For CLIP, ViT-B (86M parameters) and ViT-L (0.3B parameters) also add 0.5M and 1.5M parameters from LoRA, respectively, which account for only 0.68\% and 0.51\% of the total trainable parameters.}
\rev{It is important to note that LoRA adapter weights are merged with the base model weights once training is complete, as discussed in Section \ref{sec:Fine-tuning with LoRA}. This merging reduces memory usage and allows for inference speeds that are comparable to those of the original model, eliminating the need to maintain separate layers \cite{hu2021loralowrankadaptationlarge}. As a result, the extra computational load and memory usage during training is removed during inference.}

\begin{table*}[ht]
\caption{The achieved verification performances by \rev{the} baseline model (trained from scratch) and \rev{fine-tuned} DINOv2 and CLIP on different subsets \rev{of} CASIA-WebFace. The results are reported in (\%) on the small benchmarks and as average accuracies. On IJB-B and IJB-C, the results are reported as TAR at FAR of 1e-3, 1e-4 and 1e-5. The results of the first two rows are obtained from DINOv2 and CLIP models without fine-tuning. It is worth noting that fine-tuning DINOv2 and CLIP achieved higher recognition accuracies than training Vi\rev{T} models from scratch.}
\label{table:portion_data}
\centering
\resizebox{\linewidth}{!}{%
\begin{tabular}{ccc|cccccc|cccccc}

\multirow{3}{*}{} &   & & & &  & &  &  &  &  &  &  &  & \\
                  \# Identities & \# Images & Method & LFW & CFP-FP & AgeDB30 & CALFW & CPLFW & Avg. & \multicolumn{3}{c}{IJB-B} & \multicolumn{3}{c}{IJB-C} \\
                  &   & & &   & &  &  & & $10^{-3}$ & $10^{-4}$ & $10^{-5}$ & $10^{-3}$ & $10^{-4}$ & $10^{-5}$ \\ \hline
- & - & DINOV2 &  78.73 & 71.15 & 54.70 & 59.47 & 59.48 & 64.70 & 14.66 & 5.90 & 2.33 & 18.10  & 7.44 & 2.87 \\
- & - & CLIP &   93.33 & 88.86 & 74.67 & 77.13 & 79.23  & 82.64 & 49.19 & 27.79 & 11.93 & 52.21 & 32.40 & 16.34 \\ \hline \hline

\multirow{3}{*}{$1$K} & \multirow{3}{*}{82425} 
                & Baseline  & 88.33 & 65.21 & 61.07 & 73.35 & 61.85 & 69.96 & 2.44 & 0.93 & 0.54 & 2.69  & 1.08 & 0.55 \\
                & & DINOv2  &  96.82 & 87.31 & 82.20 & 85.92 & 83.27 & 87.10 & 65.87 & 45.28 & 25.54 & 70.82 & 51.32 & 32.66\\
                & & CLIP  & 98.55 & 93.11 & 85.28 & 88.98 & 87.83 & 90.75 & 70.56 & 43.43 & 16.36 & 75.70 & 51.01 & 24.86 \\ \hline
\multirow{3}{*}{\rev{$2.5$K}} & \multirow{3}{*}{163945} 
                & Baseline  & 93.17 & 74.70 & 69.93 & 78.32 & 71.13 & 77.45 & 5.38 & 1.23 & 0.51 & 5.04 & 1.12 & 0.41 \\
                & & DINOv2  &  97.80 & 89.60 & 84.25 & 87.72 & 85.15 & 88.90 & 72.21 & 52.88 & 25.12 & 77.05 & 59.88 & 37.23 \\
                & & CLIP  & 98.87 & 93.51 & 86.12 & 90.07 & 88.78 & 91.47 & 71.03 & 44.12 & 18.42 & 76.38 & 52.58 & 26.97 \\ \hline
\multirow{3}{*}{\rev{$2.5$K} (S)} & \multirow{3}{*}{289228} 
                & Baseline  & 95.78 & 82.89 & 78.33 & 83.25 & 77.72 & 83.59 & 30.37 &  4.81 & 1.00 & 26.91 &  4.04 & 0.83\\
                & & DINOv2  &  98.50 & 90.81 & 87.25 & 88.68 & 85.93  & 90.23 & 77.28 & 61.69 & 38.96 & 80.68 & 66.97 & 49.67 \\
                & & CLIP  & 98.63 & 94.23 & 86.28 & 89.50 & 88.20 & 91.37 & 72.16  & 41.36  & 13.18 & 77.46  & 51.09 & 20.48 \\  \hline                
\multirow{3}{*}{$5$K} & \multirow{3}{*}{280215} 
                & Baseline  & 96.32 & 81.71 & 78.25 & 84.23 & 78.25 &  83.75 & 28.63 & 5.91 & 1.11 & 24.17 & 4.79 & 1.00\\
                & & DINOv2  &  98.43 & 90.81 & 86.83  &  89.17  &  85.55  & 90.16 & 76.58 & 61.34 & 38.03 & 80.81 & 66.96  & 48.81\\
                & & CLIP  & 99.13 & 94.27 & 86.85 & 90.50 & 88.87 & 91.92 & 75.65 & 53.65 & 26.92 & 80.25  & 59.21  &  35.83 \\ \hline
\multirow{3}{*}{\rev{$7.5$K}} & \multirow{3}{*}{389007} 
                & Baseline  & 97.07 & 85.71 & 81.55 & 86.20 & 81.55 & 86.42 & 56.90 & 18.05 & 3.03 & 55.15 & 17.40 & 3.80 \\
                & & DINOv2  &  98.40 & 91.94 & 87.20 & 89.63 & 85.60  & 90.55 & 74.36 &  46.42 & 15.00 & 77.96 & 52.97 & 22.29\\
                & & CLIP  & 99.13  & 94.49 & 87.33 & 90.43 & 88.45 & 91.97 & 77.92 & 56.26  & 25.84 &  81.77 & 62.78  & 34.74 \\ \hline
\multirow{3}{*}{$10$K} & \multirow{3}{*}{490623} 
                & Baseline  & 98.02 & 88.04 & 84.70 & 88.25 & 83.78 & 88.56 & 73.00 & 39.61 & 10.48 & 73.03 & 36.77 & 10.79 \\
                & & DINOv2  &  98.38 & 91.57 & 88.22 & 89.87 & 86.67 & 90.94 & 79.27  & 63.24 & 34.63 & 82.46 & 69.50 & 48.45 \\
                & & CLIP  & 98.97  & 94.29 & 87.62 & 90.62 & 89.13 & 92.13 & 80.50 & 61.45 & 33.48 & 83.96  & 67.45 & 42.45 \\

\end{tabular}}
\end{table*}

%We first evaluate the performance of foundation models on the previously mentioned benchmarks without fine-tuning. As reported in Table \ref{FM_nofine-tuning}, DINOv2 underperforms on the various benchmarks compared to CLIP. This result can be attributed to the fact that the curated dataset LVD-142M, on which DINOv2 was trained, involved a post-processing step that blurred identifiable faces \cite{oquab2024DINOv2learningrobustvisual}.
%%%%%%%%%%%%%%%%%%%%%%%%%%%%%%%%%%%%%%%%%%%%%%%%%%%%%%%%%%%%%%%%%%%%%%%%%%%%%%%%
\section{Results}
\subsection{Are Foundation Models ready for Face Recognition?} \label{sec:eval_1}
We first evaluate the face verification performances of the considered foundation models, DINOv2 and CLIP, on several challenging benchmarks described in Section \ref{sec:experimental}. For DINOv2 and CLIP, we utilized the official pre-trained models released by the corresponding authors \cite{oquab2024DINOv2learningrobustvisual, radford2021learningtransferablevisualmodels}. The models, in this case, are utilized as feature extractors without fine-tuning the model weights.

The results achieved by the different pre-trained architectures of DINOv2 and CLIP are presented in Table \ref{table:FM_nofine-tuning}. All network architectures are based on vision-transformer (Vi\rev{T}).
For details on network architectures, one can refer to the corresponding papers \cite{oquab2024DINOv2learningrobustvisual, radford2021learningtransferablevisualmodels}. 
The results of the first row (noted as baseline (MS1MV2)) are achieved by a Vi\rev{T}-S model trained from scratch on MS1MV2. These evaluation results are provided in this table as a reference.
One can clearly observe that different pre-trained CLIP model versions outperformed DINOv2 models on the considered benchmarks. This result might be attributed to the fact that the curated dataset LVD-142M \cite{oquab2024DINOv2learningrobustvisual}, on which DINOv2 was trained, involved a post-processing step that blurred identifiable faces \cite{oquab2024DINOv2learningrobustvisual}.
Although the considered foundation models are not trained and optimized to perform feature extraction for face verification, they achieved relatively high accuracies on \rev{the} considered benchmarks. For example, the achieved verification accuracy by CLIP (ViT-L/14@336), the largest model from CLIP, was $96.30\%$ on LFW. On the benchmarks with cross-age evaluation protocol, the best-achieved accuracies on AgeDB30 and CALLFW were $79.80\%$ and $81.83\%$, respectively. All considered models achieved relatively low TAR at the reported thresholds of FAR on large and challenging benchmarks, IJB-B and IJB-C. 

From the reported results in Table \ref{table:FM_nofine-tuning}, one can conclude that the achieved results by foundation models on face verification are far from being random. They achieved relatively high accuracies on less challenging benchmarks (LFW). However, this performance significantly drops when the evaluation benchmark contains challenging pairs such as AgeDB30, CALFW, IJB-B, and IJB-C, especially when considering the baseline model trained from scratch and the current performances of SOTA FR solutions \cite{Deng_2022,wang2018cosfacelargemargincosine,ElasticFace,TransFace}.

%%%%%%%%%%%%%%%%%%%%%%%%%%%%%%%%%%%%%%%%%%%%%%%%%%%%%%%%%%%%%%%%%%%%%%%%%%%%%%%%
\subsection{FRoundation: Fine-tuning Foundation Models} \label{sec:eval_2}

\begin{figure}[ht!]
  \centering
   \includegraphics[width=0.7\linewidth]{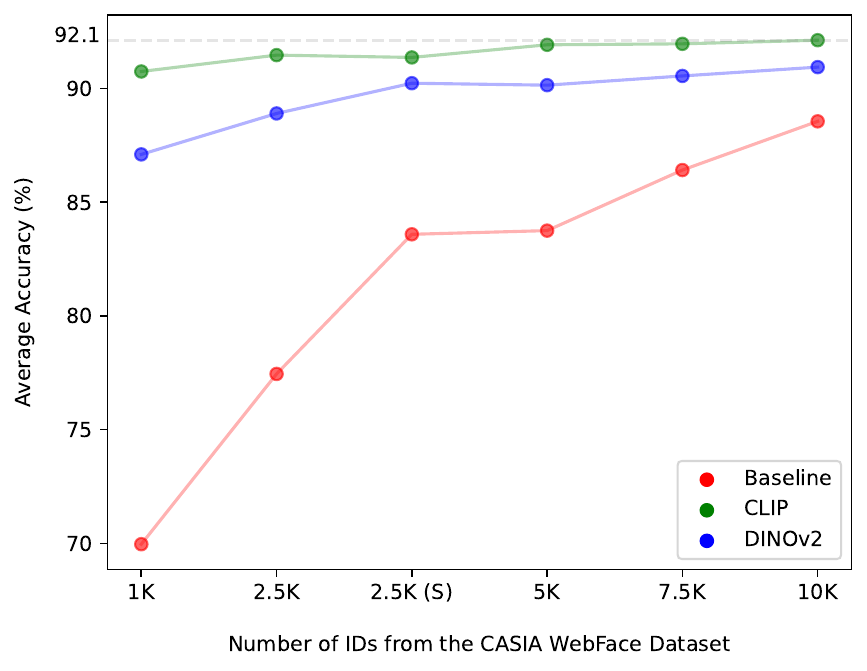}
   \caption{
    Average verification accuracies on five benchmarks, LFW, CFP-FP, AgeDB30, CALFW, and CPLFW on the y-axis vs. training/fine-tuning dataset size, in terms of the number of identities, on the x-axis. The results correspond to the ones reported in Table \ref{table:portion_data} and plotted for ViT (baseline) as well as fine-tuned DINOv2 and CLIP. Increasing the training/fine-tuning dataset width (number of identities) improved the model recognition performances.}
   \label{fig:portion_data}
\end{figure}

Table \ref{table:portion_data} presents the verification accuracies achieved by considered foundation models, DINOv2 and CLIP, fine-tuned with LoRA (Section \ref{sec:FRoundation}) on different subsets of CASIA-WebFace. The results noted as the baseline are achieved by ViT-S trained from scratch and provided as a reference. 
In this experiment, we utilized the smallest released pre-trained ViT architecture from DINOv2 (ViT-S) and CLIP (ViT-B). 
The results of the first two rows in Table \ref{table:portion_data} are achieved by pre-trained DINOv2 (ViT-S) and CLIP (ViT-B) without fine-tuning.
%Given that the pre-trained foundation models achieved relatively high accuracies on several benchmarks (Table \ref{table:FM_nofine-tuning}) and 
Driven by possible technical limitations and ethical concerns in practice to collect large-scale face (or other biometric) datasets \cite{DBLP:journals/ivc/BoutrosSFD23}, we propose to fine-tune the foundation models on small subsets from CAISA-WebFace and compare them to the case where the model was trained from scratch. Specifically, we provided a border evaluation of the impact of dataset size in terms of number of identities (dataset width) on the model performance by utilizing $(n)$K identities from \rev{CASIA}-WebFace, where $n \in [1, 2.5, 5, 7.5, 10]$. 
Additionally, we investigate the impact of the dataset depth (number of images) on the model performance by comparing the case where the $2.5$K identities are randomly selected to the case where these $2.5K$ identities are selected (noted as $2.5$K(S)) with the largest number of images per identity. The $2.5$K and $2.5$K(S) resulted in a total of $163945$ and  $289228$ images, respectively.
Based on the reported results in Table \ref{table:portion_data}, we made the following observations:
%In this section, we examine the effect of fine-tuning foundation models compared to training the ViT from scratch using small subsets of data. DINOv2 follows the ViT-S architecture, while CLIP adopts the ViT-B model, as explained in Section \ref{subsec:dataset and implementation}. To investigate the effect of the amount of data used for fine-tuning on model performance, we fine-tuned the different models on 1k, 2.5k, 5k, 7.5k, and 10k identities, with 10k representing the full CASIA WebFace dataset. Additionally, we fine-tune with the 2.5k identities that have the largest number of images per identity, referred to as 2.5k selective (S).
%Table \ref{table:portion_data} presents the verification accuracies of the foundation models fine-tuned on different subsets of CASIA-WebFace. We can make the following observations:
\begin{itemize}
    \item Fine-tuning DINOv2 and CLIP on different subsets from CASIA-WebFace significantly improved the achieved verification accuracies on all evaluation benchmarks, in comparison to the case where DINOv2 and CLIP are utilized without fine-tuning. Initially, without fine-tuning, DINOv2 (ViT-S) and CLIP (ViT-B) achieved average accuracies of $64.70$\% and $82.64$\%, respectively. These average accuracies increased to $87.10\%$ and $90.75\%$ by fine-tuning DINOv2 and CLIP, respectively\rev{,} on a small subset of only 1\rev{K} identities from CAISA-WebFace.
    \item As expected, using a larger subset for fine-tuning DINOv2 and CLIP consistently resulted in higher verification accuracies across all considered benchmarks. The same observation can also be made for the model trained from scratch (baseline). 
    \item \rev{F}ine-tuning ViT of pre-trained DINOv2 and CLIP led to higher verification accuracies, in comparison to the case where the model is trained from scratch. The superiority of foundation models over the model trained from scratch on CASIA-WebFace (or a subset of CASIA-WebFace) can be observed on the small evaluation benchmarks and the large-scale, IJB-B and IJB-C, benchmarks. This observation can be seen also in Figure \ref{fig:portion_data}.
    \item Figure \ref{fig:portion_data} presents average accuracies  (y-axis) of baseline and foundation models fine-tuned on different subsets (x-axis) from CASIA-WebFace. One can notice that the average accuracies slightly improved when a larger subset of CASIA-WebFace is utilized. On the other hand, the performance of the baseline significantly increases with more data, starting at an average accuracy of $69.96$\% when trained on $1K$ and reaching $88.56$\% on the full dataset with an increase of $18.6$\%. The same observation can also be made \rev{for the IJB-B and IJB-C benchmarks} (Table \ref{table:portion_data}).
    \item  The impact of the dataset depth (number of images) can be observed by comparing the achieved results of models trained/\rev{fine-tuned} on $2.5$K ($163945$) and $2.5$K (S) ($289228$). One can \rev{observe} that increasing the dataset depth leads to higher verification accuracies in most settings.

\end{itemize}

%%%%%%%%%%%%%%%%%%%%%%%%%%%%%%%%%%%%%%%%%%%%%%%%%%%%%%%%%%%%%%%%%%%%%%%%%%%%%%%%
\subsection{Large-scale Fine-tuning}
 
\begin{table*}[ht]
 \caption{\rev{The achieved verification accuracies by baseline models, as well as fine-tuned DINOv2 and CLIP, are presented on several challenging benchmarks}. Different architectures of DINOv2 and CLIP are fine-tuned on different datasets, CASIA-WebFace, MS1MV2, and WebFace4M. The results of the baseline refer to models trained from scratch.}
 \label{tab:full_training}
\resizebox{\linewidth}{!}{%
 \centering
 \begin{tabular}{ccc|cccccc|cccccc}

 \multirow{3}{*}{} &  & & & & &  &  &  &  &  &  \\
                   Method & Backbone & Train data & LFW & CFP-FP & AgeDB30 & CALFW & CPLFW & Avg. &\multicolumn{3}{c}{IJB-B} & \multicolumn{3}{c}{IJB-C} \\
                   &  & &  & &  & & & & $10^{-3}$ & $10^{-4}$ & $10^{-5}$ & $10^{-3}$ & $10^{-4}$ & $10^{-5}$ \\ \hline
\multirow{6}{*}{Baseline} 
                 & \multirow{3}{*}{ViT-S} & CASIA-WebFace & 98.02 & 88.04 & 84.70 & 88.25 & 83.78 & 88.56 & 73.00 & 39.61 & 10.48 & 73.03 & 36.77 & 10.79 \\
                 &  & MS1MV2 & 99.73 & 95.44 &  97.42 & 95.95 & 92.35 & 96.18 & 95.89 & 92.85 & 79.68 & 96.76 &  94.51 & 88.30\\ 
                 &  & WebFace4M & 99.63 & 96.57 & 96.20 & 95.55 & 92.92 & 96.17 & 96.34 & 93.86  & 88.30 & 97.44 & 95.62 & 92.46 \\  [5pt]
                 & \multirow{3}{*}{ViT-L} & CASIA-WebFace & 97.87 & 87.84 & 81.53 & 87.03 & 82.67 & 87.39 & 76.45 & 58.96 & 37.88 & 79.26 & 62.51 & 43.95 \\ 
                 &  & MS1MV2 & 99.73 & 95.31 & 96.48 & 95.68 & 92.22 & 95.88 & 94.76 & 90.07 & 74.38 & 95.85 & 92.49 & 82.71 \\ 
                 &  & WebFace4M & 99.68 & 96.47 & 94.60 & 94.85 & 92.63 & 95.65 & 95.82 & 92.92 & 85.94 & 97.14 & 94.79 & 90.90\\ \hline 

 \multirow{6}{*}{DINOv2} 
                 & \multirow{3}{*}{ViT-S} & CASIA-WebFace &  98.38 & 91.57 & 88.22 & 89.87 & 86.67 & 90.94 & 79.27  & 63.24 & 34.63 & 82.46 & 69.50 & 48.45 \\ 
                 &  & MS1MV2 & 99.02 & 89.71 & 91.42 & 92.90 & 86.83 & 91.98 & 89.52 & 81.70 & 69.10 & 91.61  & 85.25 & 76.95 \\ 
                 &  & WebFace4M & 98.95 & 91.43 & 87.77 & 91.43 & 87.73 & 91.46 & 89.71 & 81.13 & 68.34 &  92.08 & 85.13 & 76.10 \\  [5pt]
                 & \multirow{3}{*}{ViT-L} & CASIA-WebFace & 99.33 & 95.77 & 92.77 & 92.33 & 91.20 & 94.28 & 89.98 & 78.81  & 61.77  & 92.44 & 84.45 & 72.91 \\ 
                 &  & MS1MV2 & 99.63 & 95.93 & 96.22 & 95.50 & 92.78 & 96.01 & 95.31 & 91.95 & 83.50 & 96.41  & 93.94 & 89.89 \\ 
                 &  & WebFace4M & 99.65 & 96.84 & 95.10 & 94.80 & 93.75 & 96.03 & 95.64 & 92.65 & 86.42  & 96.96 & 94.79 & 91.40 \\  \hline 
                 
 \multirow{6}{*}{CLIP} 
                 & \multirow{3}{*}{ViT-B} & CASIA-WebFace & 98.97  & 94.29 & 87.62 & 90.62 & 89.13 & 92.13 & 80.50 & 61.45 & 33.48 & 83.96  & 67.45 & 42.45 \\
                 &  & MS1MV2 & 99.43 & 93.51 & 92.02 & 93.37 & 90.40 & 93.75 & 90.82 & 82.39 & 65.17 & 92.82 & 86.31 & 75.32 \\   
                 &  & WebFace4M & 99.30 & 93.93 & 88.90 & 92.75 & 90.67 & 93.11 & 90.71 & 81.52 & 68.05 & 92.85 & 85.63 & 75.73 \\ [5pt]
                 & \multirow{3}{*}{ViT-L} & CASIA-WebFace &  99.55 & 95.73 & 91.73 & 92.58 & 91.70 & 94.26 & 90.33 & 78.71 & 56.96 & 92.59 & 83.12 & 68.87 \\
                 &  & MS1MV2 &  99.68 & 96.76 & 93.20 & 94.60 & 93.73  & 95.59 & 95.73  & 91.22 & 82.78 & 96.80 & 93.66 & 88.62 \\  
                 &  & WebFace4M & 99.65 & 96.50 & 93.72 & 94.37 & 93.73 & 95.59 & 95.12 & 90.72 & 82.76  & 96.62 & 93.40  &  88.55  \\

 \end{tabular}}
 \end{table*}

\begin{table*}[!t]
\caption{\rev{The achieved verification accuracies by notable previous works, namely CosFace, ArcFace, CurricularFace, MagFace, ElasticFace, and AdaFace, are reported on IJB-B and IJB-C as TAR at a FAR of 1e-4, along with various small benchmarks.}}
\label{tab:literature_results}
\resizebox{\linewidth}{!}{%
  \centering
\begin{tabular}{cc|cccccc|cc}
            Method  & Train data & LFW & CFP-FP & AgeDB30 & CALFW & CPLFW & Avg. & IJB-B & IJB-C \\ \hline
            \rev{CosFace} \cite{wang2018cosfacelargemargincosine, zhu2021webface260mbenchmarkunveilingpower} \rev{(CVPR 2018)} & \rev{MS1MV2} & \rev{99.81} & \rev{98.18} & \rev{98.34} & \rev{96.18} & \rev{92.76} & \rev{97.05} & \rev{-} & \rev{96.01} \\
            \rev{ArcFace} \cite{Deng_2022, zhu2021webface260mbenchmarkunveilingpower} \rev{(CVPR2019)} & \rev{MS1MV2} & \rev{99.78} & \rev{98.54} & \rev{98.21} & \rev{96.05} & \rev{92.93} & \rev{97.10} &  \rev{-} & \rev{96.03} \\
            \rev{CurricularFace} \cite{huang2020curricularfaceadaptivecurriculumlearning, zhu2021webface260mbenchmarkunveilingpower} \rev{(CVPR2020)} &  \rev{MS1MV2}  & \rev{99.83} & \rev{98.67} & \rev{98.32} & \rev{96.28} & \rev{93.05} & \rev{97.23} &  \rev{-} & \rev{96.21 } \\
            \rev{MagFace} \cite{meng2021magfaceuniversalrepresentationface} \rev{(CVPR2021)} & \rev{MS1MV2}  & \rev{99.83} & \rev{98.46} & \rev{98.17} & \rev{96.15} & \rev{92.87} & \rev{97.09} &  \rev{94.51} & \rev{95.97} \\
            \rev{ElasticFace-Arc} \cite{ElasticFace} \rev{(CVPRW2022)} &  \rev{MS1MV2}  & \rev{99.80} & \rev{98.67} & \rev{98.35} & \rev{96.17} & \rev{93.27} & \rev{97.25} &  \rev{95.22} & \rev{96.49} \\
            \rev{AdaFace} \cite{kim2023adafacequalityadaptivemargin} \rev{(CVPR2022)} &  \rev{MS1MV2}  & \rev{99.82} & \rev{98.49} & \rev{98.05} & \rev{96.08} & \rev{93.53} & \rev{97.19} &  \rev{95.67} & \rev{96.89} \\ \hline
            
            \rev{CosFace} \cite{wang2018cosfacelargemargincosine, zhu2021webface260mbenchmarkunveilingpower} \rev{(CVPR 2018)} &  \rev{WebFace4M}  & \rev{99.80} & \rev{99.25} & \rev{97.45} & \rev{95.95} & \rev{94.40} & \rev{97.37} &  \rev{-} & \rev{96.86} \\
            \rev{ArcFace} \cite{Deng_2022, zhu2021webface260mbenchmarkunveilingpower} \rev{(CVPR2019)} & \rev{WebFace4M}  & \rev{99.85} & \rev{99.04} & \rev{97.82} & \rev{95.93} & \rev{94.31} & \rev{97.39} &  \rev{-} & \rev{96.77} \\
            \rev{CurricularFace} \cite{huang2020curricularfaceadaptivecurriculumlearning, zhu2021webface260mbenchmarkunveilingpower} \rev{(CVPR2020)} &  \rev{WebFace4M}  & \rev{99.83} & \rev{99.11} & \rev{97.83} & \rev{96.03} & \rev{94.21} & \rev{97.40} &  \rev{-} & \rev{97.02 } \\
            \rev{AdaFace} \cite{kim2023adafacequalityadaptivemargin} \rev{(CVPR2022)} &  \rev{WebFace4M}  & \rev{99.80} & \rev{99.17} & \rev{97.90} & \rev{96.05} & \rev{94.63} & \rev{97.51} &  \rev{96.03} & \rev{97.39} \\ 
\end{tabular}}
\end{table*}

Table \ref{tab:full_training} presents the achieved recognition performances by models trained from scratch (baseline) and fine-tuned (CLIP and DINOV2) on a relatively small dataset, CASIA-WebFace ($0.5$\rev{M} images) and larger datasets, MS1MV2 ($5.8$\rev{M} images) and \rev{WebFace4M} ($4$\rev{M} images). All results are reported using large (ViT-L) and small (ViT-S of DINOv2 and baseline and ViT-B of CLIP) architectures.
We made the following observation from the reported results in Table \ref{tab:full_training}:
\begin{itemize}
    \item Using CASIA-WebFace ($0.5$\rev{M} images), fine-tuning DINOv2 (ViT-S and ViT-L architectures) and CLIP  (ViT-B and ViT-L architectures) led to higher verification accuracies, in comparison to the case where ViT-S and ViT-L are trained from scratch on the same data.  For example, the average verification accuracies on the small benchmarks, LFW, CFP-FP, AgeDB30, CALFW, and CPLFW, was $87.39$\% by baseline (ViT-L) which is lower than $94.28$\% and $94.26$\% achieved by the \rev{fine-tuned} DINOv2 and CLIP, respectively.  
    \item The models that are trained/\rev{fine-tuned} on large datasets,  MS1MV2 and \rev{WebFace4M}, achieved higher verification accuracies than the one trained on the relatively smaller dataset, CASIA-WebFace.
    \item Using small architectures (ViT-S and ViT-B) and large-scale training datasets (MS1MV2 or WebFace4M), the model trained from scratch achieved higher verification accuracies than the \rev{fine-tuned} DINOv2 and CLIP. This observation can be noticed from the achieved results on small benchmarks as well as on large-scale, IJB-B and IJB-C, benchmarks. 
    \item Using large architectures (ViT-L) and large-scale training/fine-tuning datasets (MS1MV2 or WebFace4M), the \rev{fine-tuned} DINOv2 and CLIP achieved slightly higher recognition performance than the models trained from scratch on most of the evaluation benchmarks.  
\end{itemize}

To conclude, for the case where only a relatively small training dataset is accessible, fine-tuning pre-trained foundation models can achieve higher recognition accuracy than training the same model architecture from scratch.  In the case that one has access to large training datasets, training a model from scratch for FR can achieve very competitive results to fine-tuning foundation models. However, this, next to the technical and legal limitations of collecting or maintaining the data, comes with a high training time cost. For example, training ViT-L from scratch on MS1MV2 using the settings described in Section \ref{sec:experimental} requires around $70$ GPU hours, in comparison to around $40$ GPU hours for fine-tuning DINOv2 with LoRA on $8$ Nvidia A100 SXM4 $40$GB GPUs. \rev{Additionally, to put the performances achieved by FRoundation in the context of some of the major works in FR, we present in Table \ref{tab:literature_results}  results of the most influential works in the field \cite{wang2018cosfacelargemargincosine, Deng_2022, meng2021magfaceuniversalrepresentationface, ElasticFace, huang2020curricularfaceadaptivecurriculumlearning, kim2023adafacequalityadaptivemargin}. However, this comparison is not direct, as such works commonly use the ResNet-100 \cite{He2015DeepRL} architecture and the MS1MV2 or WebFace4M dataset \cite{Deng_2022, zhu2021webface260mbenchmarkunveilingpower} for training.} \rev{The various solutions outperform the trained from scratch models presented in Table \ref{tab:full_training}. However, the results achieved by our FRoundation enhance the performance to a very close level to the works of the solutions presented in table \ref{tab:literature_results}.}

%%%%%%%%%%%%%%%%%%%%%%%%%%%%%%%%%%%%%%%%%%%%%%%%%%%%%%%%%%%%%%%%%%%%%%%%%%%%%%%%

\begin{table*}[htbp]
\caption{\rev{The verification accuracies obtained by the baseline (trained from scratch) models, along with the foundation models DINOv2 and CLIP, are presented after training/fine-tuning on various synthetic datasets, including IDiff-Face and SFace2.}}
\label{tab:training_synthetic}
\resizebox{\linewidth}{!}{%
 \centering
 \begin{tabular}{ccc|cccccc|cccccc}
\multirow{3}{*}{} &  & & & & &  &  &  &  &  &  \\
                   Method & Backbone & Train data & LFW & CFP-FP & AgeDB30 & CALFW & CPLFW & Avg. &\multicolumn{3}{c}{IJB-B} & \multicolumn{3}{c}{IJB-C} \\
                   &  & &  & &  & & & & $10^{-3}$ & $10^{-4}$ & $10^{-5}$ & $10^{-3}$ & $10^{-4}$ & $10^{-5}$ \\ \hline
\multirow{4}{*}{\rev{Baseline}} 
                 & \multirow{2}{*}{\rev{ViT-S}} 
                 & \rev{SFace2} & \rev{92.70} & \rev{69.84} & \rev{69.42} & \rev{78.77} & \rev{65.87} & \rev{75.32} & \rev{13.51} & \rev{4.30} & \rev{1.79} & \rev{11.71} & \rev{3.72} & \rev{1.47} \\ 
                 & & \rev{IDiff-Face} & \rev{96.35} & \rev{77.50} & \rev{79.48} & \rev{85.78} & \rev{75.57} & \rev{82.93} & \rev{60.30} & \rev{33.77} & \rev{12.55} & \rev{58.33} & \rev{30.08} & \rev{11.61} \\ [5pt]
                 & \multirow{2}{*}{\rev{ViT-L}} 
                 & \rev{SFace2} & \rev{91.37} & \rev{71.56} & \rev{68.05} & \rev{75.83} & \rev{68.82} & \rev{75.13} & \rev{40.94} & \rev{18.65} & \rev{6.62} & \rev{36.31} & \rev{14.53} & \rev{5.34} \\ 
                 &  & \rev{IDiff-Face} & \rev{96.13} & \rev{76.56} & \rev{77.43} & \rev{85.25} & \rev{75.83} & \rev{82.24} & \rev{56.93} & \rev{26.14} & \rev{6.07} & \rev{54.30} & \rev{21.20} &  \rev{4.20} \\  \hline 
\multirow{6}{*}{\rev{DINOv2}} 
                 & \multirow{3}{*}{\rev{ViT-S}} 
                 & \rev{-} & \rev{78.73} & \rev{71.15} & \rev{54.70} & \rev{59.47} & \rev{59.48} & \rev{64.70} & \rev{14.66} & \rev{5.90} & \rev{2.33} & \rev{18.10} & \rev{7.44} & \rev{2.87} \\
                 &  & \rev{SFace2} & \rev{93.57} & \rev{76.97} & \rev{73.70} & \rev{78.95} & \rev{71.85} & \rev{79.01} & \rev{58.14} & \rev{39.84} & \rev{22.77} & \rev{60.12} & \rev{43.23} & \rev{28.00} \\ 
                 &  & \rev{IDiff-Face} & \rev{96.13} & \rev{80.54} & \rev{80.13} & \rev{86.12} & \rev{74.32} & \rev{83.45} & \rev{64.86} & \rev{42.01} & \rev{15.99} & \rev{67.23} & \rev{47.80} & \rev{23.58} \\ [5pt]  
                 
                 & \multirow{3}{*}{\rev{ViT-L}} 
                 & \rev{-} & \rev{80.37} & \rev{71.97} & \rev{55.25} & \rev{60.52} & \rev{64.33} & \rev{66.49} & \rev{17.18} & \rev{6.44} & \rev{2.52} & \rev{20.36} & \rev{7.84} & \rev{2.77} \\
                 &  & \rev{SFace2} & \rev{95.55} & \rev{84.58} & \rev{77.73} & \rev{82.48} & \rev{78.10} & \rev{83.69} & \rev{68.33} & \rev{48.92} & \rev{28.30} &  \rev{71.37} & \rev{53.32}  & \rev{34.94} \\  
                 &  & \rev{IDiff-Face} & \rev{97.77} & \rev{86.94} & \rev{86.67} & \rev{90.28} & \rev{81.73} & \rev{88.68} & \rev{75.11} & \rev{55.75} & \rev{32.80} & \rev{78.75} & \rev{62.17} & \rev{40.94} \\ \hline 
\multirow{6}{*}{\rev{CLIP}} 
                 & \multirow{3}{*}{\rev{ViT-B}} 
                 & \rev{-} & \rev{93.33} & \rev{88.86} & \rev{74.67} & \rev{77.13} & \rev{79.23}  & \rev{82.64} & \rev{49.19} & \rev{27.79} & \rev{11.93} & \rev{52.21} & \rev{32.40} & \rev{16.34} \\
                 &  & \rev{SFace2} & \rev{93.98} & \rev{86.56} & \rev{76.77} & \rev{81.95} & \rev{78.45} & \rev{83.54} & \rev{66.07} & \rev{39.93} & \rev{14.12} & \rev{69.44} & \rev{47.24} & \rev{23.13} \\
                 &  & \rev{IDiff-Face} & \rev{97.58} & \rev{87.91} & \rev{80.78} & \rev{88.17} & \rev{80.48} & \rev{86.98} & \rev{74.47} & \rev{54.15}  & \rev{26.27} & \rev{77.54} & \rev{61.01} & \rev{37.43} \\  [5pt] 
                 
                 & \multirow{3}{*}{\rev{ViT-L}} 
                 & \rev{-}  & \rev{95.90} & \rev{90.66} & \rev{79.82} & \rev{83.10} & \rev{82.73}  & \rev{86.44} & \rev{62.72} & \rev{40.90} & \rev{20.52} & \rev{64.74} & \rev{44.69} & \rev{25.23} \\
                 &  & \rev{SFace2} & \rev{97.88} & \rev{89.87} & \rev{79.03} & \rev{86.50} & \rev{83.88} & \rev{87.43} & \rev{75.31} & \rev{54.86} & \rev{25.14} & \rev{78.64} & \rev{60.81} & \rev{37.73} \\
                 &  & \rev{IDiff-Face} & \rev{98.75} & \rev{91.40} & \rev{86.50} & \rev{90.78} & \rev{84.87} & \rev{90.46} & \rev{82.92} & \rev{64.56} & \rev{34.82} & \rev{86.07} & \rev{71.77} & \rev{48.18} \\
 \end{tabular}}
 \end{table*}

\subsection{\rev{Learning from Synthetic Data}}
\rev{Table \ref{tab:training_synthetic} presents the verification accuracies obtained by models trained from scratch (baseline) on IDiff-Face \cite{boutros2023idifffacesyntheticbasedfacerecognition} and SFace2 \cite{DBLP:journals/tbbis/BoutrosHLSD24} alongside the considered foundation models, DINOv2 and CLIP, which were fine-tuned using LoRA on IDiff-Face and SFace2. All results are reported using the experimental setups presented in Section \ref{sec:experimental}. From the results presented in Table \ref{tab:training_synthetic}, we draw the following insights:} 

\begin{itemize}
    %\item The baseline models trained on IDiff-Face outperformed models trained on SFace2 on all evaluated benchmarks.
   % \item \rev{Baseline models, on most of the evaluated benchmarks, underperform compared to fine-tuned models, DINOv2 and CLIP. Indicating that pre-training significantly enhances model accuracy.}
    \item \rev{The recognition performance of CLIP and DINOv2 improves when fine-tuning with any of the considered synthetic datasets compared to using the pre-trained models alone, highlighting the significant potential of synthetic datasets in enhancing model performance.}
    \item \rev{The models trained/fine-tuned on the IDiff-Face dataset generally deliver higher recognition performance across various benchmarks compared to those trained on the SFace2 dataset. For example, the ViT-L model of CLIP achieves higher verification accuracy on all evaluation benchmarks when fine-tuned on IDiff-Face, compared to when it is fine-tuned on SFace2. On the small benchmarks, the model achieved an average accuracy of 90.46\% when fine-tuned on IDiff-Face, compared to 87.43\% when fine-tuned on SFace2. A similar trend is observed on the large benchmarks, where the model fine-tuned on IDiff-Face achieved accuracies of 64.56\% and 71.77\% on IJB-B and IJB-C at a FAR of 1e-4, respectively, while the model fine-tuned on SFace2 achieved accuracies of 54.86\% and 60.81\% on the same benchmarks.}
    \item \rev{Fine-tuning the larger ViT models of pre-trained DINOv2 and CLIP consistently led to higher verification accuracies on most benchmarks compared to their smaller counterparts. This is not the case for the baseline models, where both the ViT-S and ViT-L demonstrate competitive performance across various benchmarks. This can be attributed to the fact that larger models are more prone to overfitting when trained on smaller-scale datasets \cite{dosovitskiy2021imageworth16x16words}.}
\end{itemize}

\rev{This investigation highlights the potential of synthetic data in enhancing the performance of foundation models for FR, demonstrating superior results compared to both pre-trained and baseline models. Furthermore, it encourages further exploration of additional synthetic datasets for FR, such as DCFace \cite{kim2023dcfacesyntheticfacegeneration} and the more recent ID$^{3}$ \cite{li2024id3identitypreservingyetdiversifieddiffusionmodels}.}

%%%%%%%%%%%%%%%%%%%%%%%%%%%%%%%%%%%%%%%%%%%%%%%%%%%%%%%%%%%%%%%%%%%%%%%%%%%%%%%%
\subsection{Bias evaluations foundation models FR}

\begin{table*}[!htbp]
 \caption{Evaluation results on RFW reported as average recognition performance in (\%), standard deviation (STD) and skewed error ratio (SER) across four different demographic groups. The higher STD indicates a more biased model and the higher average (avg) indicates, in general, better recognition performance. \rev{The rightmost columns represent the average verification performance on small benchmarks (SB) and on IJB-C as TAR at FAR 1e-4, previously presented in tables \ref{table:FM_nofine-tuning} and \ref{tab:full_training}.}}
 \label{tab:RFW}
 \resizebox{\linewidth}{!}{%
 \scriptsize
 \centering
 \begin{tabular}{ccc|cccc|ccc|cc}

 \multirow{3}{*}{} & & & & & & & & & & &  \\
                   Method & Backbone & Train data & African & Asian & Caucasian & Indian & Avg. & STD & SER & Avg. & IJB-C \\
                   & & & & & & & & & & SB & 1e-4 \\ \hline

\multirow{6}{*}{Baseline} 
                 & \multirow{3}{*}{ViT-S} 
                 & CASIA-WebFace & 75.25 & 75.68 & 84.75 & 78.58 & 78.57 & 4.38 & 1.62 & \rev{88.56} & \rev{36.77} \\ 
                 &  & MS1MV2 & 96.65 & 96.32 & 98.63 & 96.68 & 97.07 & 1.05 & 2.68 & \rev{96.18} & \rev{94.51} \\ 
                 &  & WebFace4M & 94.40 & 94.12 & 97.73 & 95.02 & 95.32 & 1.65 & 2.59 & \rev{96.17} & \rev{95.62} \\   [5pt]
                 & \multirow{3}{*}{ViT-L} & CASIA-WebFace & 71.87 & 73.47 & 81.08 & 74.23 & 75.16 & 4.06 &  1.48 & \rev{87.39} & \rev{62.51} \\ 
                 &  & MS1MV2 & 97.32 & 96.48  & 98.45 & 97.25 & 97.38 & 0.81 &  2.27 & \rev{95.88} & \rev{92.49} \\ 
                 &  & WebFace4M & 93.37 & 92.25 & 96.33 & 93.27 & 93.80 & 1.75 & 2.11 & \rev{95.65}  & \rev{94.79} \\ \hline 

 \multirow{8}{*}{DINOv2} 
                 & \multirow{4}{*}{ViT-S} 
                 & - & 54.77 & 61.13 & 65.00 & 60.42 & 60.33 &  4.21 & 1.29 & \rev{64.70} & \rev{7.44} \\ 
                 &  & CASIA-WebFace & 76.15 & 76.90 & 85.98 & 80.65 & 79.92 &  4.49 & 1.70 & \rev{90.94} & \rev{69.50} \\ 
                 &  & MS1MV2 & 83.43 & 83.77 & 91.18 & 87.05 & 86.36 &  3.60 & 1.87 & \rev{91.98} & \rev{85.25}  \\ 
                 &  & WebFace4M & 80.83 & 82.90 & 88.50 & 84.77 & 84.25 &  3.25 & 1.66 & \rev{91.46} & \rev{85.13} \\   [5pt]
                 & \multirow{4}{*}{ViT-L} 
                 & - & 58.46 & 64.20 & 67.47 & 60.93 & 62.77 &  3.91 & 1.27 & \rev{66.49} & \rev{7.84}  \\ 
                 &  & CASIA-WebFace & 85.97 & 84.00 & 93.15 & 86.65 & 87.44 & 3.96 & 2.33 & \rev{94.28} & \rev{84.45} \\ 
                 &  & MS1MV2 & 93.75 & 93.40 & 97.08 & 93.92 & 94.54 & 1.70 & 2.26 & \rev{96.01} & \rev{93.94} \\ 
                 &  & WebFace4M & 92.85 & 91.75 & 95.97 & 93.40 & 93.49 & 1.78 & 2.04 & \rev{96.03} & \rev{94.79}  \\ \hline 

 \multirow{8}{*}{CLIP} 
                 & \multirow{4}{*}{ViT-B} 
                 & - & 70.75 & 69.73 & 79.32 & 68.98 & 72.19 & 4.80 & 1.49 & \rev{82.64} & \rev{32.40} \\ 
                 &  & CASIA-WebFace & 80.13 & 80.53 & 89.18 & 80.30 & 82.54 & 4.43 & 1.83 & \rev{92.13} & \rev{67.45} \\ 
                 &  & MS1MV2 & 85.60 & 86.30 & 91.82 & 87.40 & 87.78 & 2.79 & 1.76 & \rev{93.75} & \rev{86.31} \\ 
                 &  & WebFace4M & 84.43 & 84.62 & 90.80 & 85.97 & 86.45 & 2.97 & 1.69 & \rev{93.11} & \rev{85.63} \\   [5pt]
                 & \multirow{4}{*}{ViT-L} 
                 & - & 74.03 & 72.15 & 82.60 & 73.15 & 75.48 &  4.80 & 1.60 & \rev{86.44} & \rev{44.69} \\ 
                 &  & CASIA-WebFace & 84.65 & 84.47 & 92.60 & 85.02 & 86.69 & 3.94 & 2.09 & \rev{94.26} & \rev{83.12} \\ 
                 &  & MS1MV2 & 90.63 & 90.77 & 95.03 & 91.92 & 92.09 & 2.04 & 1.88 & \rev{95.59} & \rev{93.66} \\ 
                 &  & WebFace4M & 90.40 & 90.28 & 94.73 & 90.90 & 91.57 & 2.11 & 1.84 & \rev{95.59} & \rev{93.40} \\   

 \end{tabular}}
 \end{table*}

We evaluated the baseline model, CLIP, and DINOv2 on the Racial Face in the Wild (RFW) dataset \cite{RFW} to assess the models' bias and their performance across different demographic groups. The RFW dataset contains four testing subsets corresponding to\rev{:}Caucasian, Asian, Indian, and African demographic groups. Following \cite{PAMI_BIAS,RFW}, we reported the results as verification accuracies in (\%) on each subset and as average accuracies to evaluate general recognition performance. To evaluate the bias, we reported the standard deviation (STD) between all subsets and the skewed error ratio (SER), which is given by

\begin{equation} 
\begin{split}
\frac{max_gError_g}{min_gError_g}
\end{split}
\label{eq_ser}
\end{equation}

, \rev{where} $g$ \rev{represents} the demographic group, as reported in \cite{PAMI_BIAS,DBLP:journals/corr/abs-1911-10692}. A higher STD value indicates more bias across demographic groups and vice versa. For SER, the model that achieved a value closer to 1 is less biased. Table \ref{tab:RFW} presents the evaluation results on RFW. As baseline models, we report the results for ViT-S and ViT-L (noted as baseline) trained from scratch on CASIA-WebFace, MS1MV2, and WebFace4M. We also reported the results of DINOv2 and CLIP without fine-tuning and for the case where the models are fine-tuned on CASIA-WebFace, MS1MV2, or WebFace4M. One can observe the following from the reported results in Table \ref{tab:RFW}:
 \begin{itemize} 
\item \rev{Fine-tuning} DINOv2 and CLIP improved the general recognition performance on all demographics, in comparison to the case where pre-trained DINOv2 and CLIP are used without fine-tuning. These results are complementary to the ones reported in the previous section. 
\item \rev{In general, the presented models achieved unequal verification performances on different subsets of RFW, where all models achieved their best verification performances on the Caucasian subset. This can be observed from the high STD values (far from optimal zero) and SER values that are far from 1. 
These observations are aligned with the previous works \cite{PAMI_BIAS, DBLP:journals/corr/abs-1911-10692, DBLP:journals/inffus/MelziTVKRLDMFOZZYZWLTKZDBVGFFMUG24, DBLP:conf/cvpr/Gong0021, RFW}, reporting bias in FR when trained/finetuned on unbalanced datasets.
Potential causes of this bias \cite{Yucer2023RacialBW} link to the unbalanced training datasets, model's sensitivity to skin color \cite{9001031} and/or hairstyles \cite{ozturk2024accuracybiasfacialhairstyle}.% that are overrepresented or underrepresented in the training datasets.
The racial distribution of the current FR datasets \cite{DBLP:journals/corr/YiLLL14a, zhu2021webface260mbenchmarkunveilingpower, guo2016ms} used in the literature is not balanced with a majority of identities being Caucasians \cite{PAMI_BIAS,DBLP:journals/corr/abs-1911-10692, RFW}. For example, as reported in \cite{RFW}, CASIA-WebFace \cite{DBLP:journals/corr/YiLLL14a} and MS-Celeb-1M \cite{guo2016ms}, contain 84.5\% and 76.3\% Caucasians, 2.6\% and 6.6 Asian, 1.6\% and 2.6\% Indian and 11.3\% and 14.5\% African, respectively. This is also true for the widely used WebFace260M \cite{zhu2021webface260mbenchmarkunveilingpower} and its subsets, such as WebFace42M and WebFace4M, where the authors reported that the majority of identities are Caucasian. The previous works \cite{PAMI_BIAS,DBLP:journals/corr/abs-1911-10692,DBLP:journals/inffus/MelziTVKRLDMFOZZYZWLTKZDBVGFFMUG24,DBLP:conf/cvpr/Gong0021, RFW} reported that bias in these datasets would reflect in the FR algorithms, leading to higher verification accuracies on Caucasians compared to other ethnicities.}

\item Using CASIA-WebFace ($0.5$\rev{M} images) for training/fine-tuning, the \rev{fine-tuned} DINOv2 and CLIP achieved higher recognition performances than the \rev{baseline model} trained from scratch. However, DINOv2 and CLIP are slightly more biased than baseline models. For example, the STD achieved by \rev{the} baseline (ViT-S) was $4.38$, which is slightly lower than the $4.49$ and $4.43$ STD achieved by \rev{fine-tuned} DINOv2 (ViT-S) and CLIP (ViT-B).
\item The models trained from scratch on large-scale datasets, MS1MV2 ($5.8$\rev{M} images) and WebFace4M ($4$\rev{M} images) \rev{achieved} higher average recognition performances and lower standard deviation (in most of the settings) than \rev{fine-tuned} CLIP and DINOv2. For example, using the ViT-L architecture trained/\rev{fine-tuned} on WebFace4M, the baseline achieved an average accuracy of $93.80$\%, in comparison to $93.43$\% and $91.57$\% achieved by DINOv2 and CLIP, respectively. Also, in this case, the baseline model achieved a lower STD ($1.74$) than DINOv2 ($1.78$) and CLIP ($2.11$). 
\item In general, \rev{fine-tuned} DINOv2 achieved higher average accuracies and lower STD than \rev{fine-tuned} CLIP. 
\item \rev{Fine-tuning} larger model architectures of pre-trained DINOv2 (ViT-L) or CLIP (ViT-L) achieved higher recognition accuracies and lower STD than fine-tuning smaller architectures: DINOv2 (ViT-S) or CLIP (ViT-B). 
\item \rev{A higher STD and an SER further away from the optimal value of 1 indicate that there is more room for improvement in achieving uniform performance across demographics.
SER values are more sensitive to slight changes in the accuracy when the error margins are small (see Equation \ref{eq_ser}). This causes models trained or fine-tuned on MS1MV2 and WebFace4M to generally have higher SER values compared to those trained or fine-tuned on CASIA-WebFace, while typically maintaining a lower STD value. This is primarily due to the higher accuracy range of the models trained on the MS1MV2 and WebFace4M datasets (resulting in lower errors), which makes the SER more sensitive to small performance differences across demographic groups.}
\item \rev{When comparing the average verification performance on small benchmarks and IJB-C between the same model trained on WebFace4M and MS1MV2, we observe that both have comparable results in most settings. However, this observation does not hold for the performance on RFW, as the models trained on MS1MV2 outperformed those trained on WebFace4M in all cases and sometimes by a significant margin. For example, the baseline ViT large shows a performance gap of approximately 4\%. This can be attributed \cite{RFW} to the identity overlap between RFW and MS1MV2, a cleaned version derived from MS-Celeb-1M \cite{guo2016ms}.}
\end{itemize}

%%%%%%%%%%%%%%%%%%%%%%%%%%%%%%%%%%%%%%%%%%%%%%%%%%%%%%%%%%%%%%%%%%%%%%%%%%%%%%%%
\section{Conclusion} % Future work and Discussion
This paper was the first to propose and investigate the use of foundation models for the task of FR. We additionally propose the \rev{adaptation} of the foundation models to this specific task under various levels of data availability. Our experiments on multiple foundation models, training datasets, and a wide range of evaluation benchmarks led to interesting conclusions, which are summarized as follows. The studied pre-trained foundation models used as feature extractors without fine-tuning, demonstrated relatively acceptable (far from random) accuracy on less challenging face verification benchmarks like LFW. However, they \rev{underperformed} on more challenging benchmarks such as AgeDB30, CALFW, IJB-B, and IJB-C. Our \rev{adaptation} of foundation models for FR showed that fine-tuning these models on even small subsets of CASIA-WebFace, e.g. $1$K identities, significantly boosts their verification accuracy across benchmarks, outperforming models trained from scratch on similar data subsets. Additionally, increasing the dataset size (dataset width) or the number of images per identity (dataset depth) enhances performance, demonstrating the scalability of foundation models with diverse data availability. The results additionally showed that, with a limited training dataset like CASIA-WebFace, fine-tuning pre-trained foundation models outperforms training models from scratch in recognition accuracy. However, when large datasets (MS1MV2 or WebFace4M) are available, training from scratch yields competitive performance, though it incurs a significantly higher training computational cost compared to fine-tuning. \rev{This highlights the importance of selecting an appropriate training strategy according to the size of the dataset.}
Our bias evaluation results indicate that fine-tuning foundation models enhance recognition performance across demographics but introduces slightly more bias compared to baseline models trained from scratch, especially on smaller fine-tuning datasets. Models trained from scratch on larger datasets (MS1MV2 and WebFace4M) achieved superior recognition accuracy and exhibited lower bias. \rev{All models achieved their best verification performances on the Caucasian subset, which may be attributed to factors such as unbalanced training datasets, sensitivity to skin color, and hairstyles that are overrepresented or underrepresented in the training datasets.} \rev{Finally, the effectiveness of synthetic data in improving the FR performance of foundation models is demonstrated, as it outperforms both pre-trained and baseline models. This also encourages further exploration of additional synthetic datasets.} 
The outcomes of this work set the stage for broader adoption of foundation models as a basis for biometric recognition, particularly in scenarios with limited data availability, being aware of the technical and legal constraints on biometric data collection and management.

\section*{Acknowledgments}
This research work has been funded by the German Federal Ministry of Education and Research and the Hessen State Ministry for Higher Education, Research and the Arts within their joint support of the National Research Center for Applied Cybersecurity ATHENE.

%%%%%%%%%%%%%%%%%%%%%%%%%%%%%%%%%%%%%%%%%%%%%%%%%%%%%%%%%%%%%%%%%%%%%%%%%%%%%%%%
{\small
\bibliographystyle{ieee_fullname}
\bibliography{main.bib}

\begin{thebibliography}{10}\itemsep=-1pt

\bibitem{aghajanyan2020intrinsicdimensionalityexplainseffectiveness}
Armen Aghajanyan, Sonal Gupta, and Luke Zettlemoyer.
\newblock Intrinsic dimensionality explains the effectiveness of language model fine-tuning.
\newblock In {\em {ACL/IJCNLP} {(1)}}, pages 7319--7328. Association for Computational Linguistics, 2021.

\bibitem{geminiteam2024geminifamilyhighlycapable}
Rohan Anil, Sebastian Borgeaud, Yonghui Wu, Jean{-}Baptiste Alayrac, Jiahui Yu, Radu Soricut, Johan Schalkwyk, Andrew~M. Dai, Anja Hauth, Katie Millican, David Silver, Slav Petrov, Melvin Johnson, Ioannis Antonoglou, Julian Schrittwieser, Amelia Glaese, Jilin Chen, Emily Pitler, Timothy~P. Lillicrap, Angeliki Lazaridou, Orhan Firat, James Molloy, Michael Isard, Paul~Ronald Barham, Tom Hennigan, Benjamin Lee, Fabio Viola, Malcolm Reynolds, Yuanzhong Xu, Ryan Doherty, Eli Collins, Clemens Meyer, Eliza Rutherford, Erica Moreira, Kareem Ayoub, Megha Goel, George Tucker, Enrique Piqueras, Maxim Krikun, Iain Barr, Nikolay Savinov, Ivo Danihelka, Becca Roelofs, Ana{\"{\i}}s White, Anders Andreassen, Tamara von Glehn, Lakshman Yagati, Mehran Kazemi, Lucas Gonzalez, Misha Khalman, Jakub Sygnowski, and et al.
\newblock Gemini: {A} family of highly capable multimodal models.
\newblock {\em CoRR}, abs/2312.11805, 2023.

\bibitem{bao2022beitbertpretrainingimage}
Hangbo Bao, Li Dong, Songhao Piao, and Furu Wei.
\newblock Beit: {BERT} pre-training of image transformers.
\newblock In {\em {ICLR}}. OpenReview.net, 2022.

\bibitem{bommasani2022opportunitiesrisksfoundationmodels}
Rishi Bommasani, Drew~A. Hudson, Ehsan Adeli, Russ~B. Altman, Simran Arora, Sydney von Arx, Michael~S. Bernstein, Jeannette Bohg, Antoine Bosselut, Emma Brunskill, Erik Brynjolfsson, Shyamal Buch, Dallas Card, Rodrigo Castellon, Niladri~S. Chatterji, Annie~S. Chen, Kathleen Creel, Jared~Quincy Davis, Dorottya Demszky, Chris Donahue, Moussa Doumbouya, Esin Durmus, Stefano Ermon, John Etchemendy, Kawin Ethayarajh, Li Fei{-}Fei, Chelsea Finn, Trevor Gale, Lauren~E. Gillespie, Karan Goel, Noah~D. Goodman, Shelby Grossman, Neel Guha, Tatsunori Hashimoto, Peter Henderson, John Hewitt, Daniel~E. Ho, Jenny Hong, Kyle Hsu, Jing Huang, Thomas Icard, Saahil Jain, Dan Jurafsky, Pratyusha Kalluri, Siddharth Karamcheti, Geoff Keeling, Fereshte Khani, Omar Khattab, Pang~Wei Koh, Mark~S. Krass, Ranjay Krishna, Rohith Kuditipudi, and et al.
\newblock On the opportunities and risks of foundation models.
\newblock {\em CoRR}, abs/2108.07258, 2021.

\bibitem{ElasticFace}
Fadi Boutros, Naser Damer, Florian Kirchbuchner, and Arjan Kuijper.
\newblock Elasticface: Elastic margin loss for deep face recognition.
\newblock In {\em {IEEE/CVF} Conference on Computer Vision and Pattern Recognition Workshops, {CVPR} Workshops 2022, New Orleans, LA, USA, June 19-20, 2022}, pages 1577--1586. {IEEE}, 2022.

\bibitem{boutros2023idifffacesyntheticbasedfacerecognition}
Fadi Boutros, Jonas~Henry Grebe, Arjan Kuijper, and Naser Damer.
\newblock Idiff-face: Synthetic-based face recognition through fizzy identity-conditioned diffusion models.
\newblock In {\em {ICCV}}, pages 19593--19604. {IEEE}, 2023.

\bibitem{DBLP:journals/tbbis/BoutrosHLSD24}
Fadi Boutros, Marco Huber, Anh~Thi Luu, Patrick Siebke, and Naser Damer.
\newblock Sface2: Synthetic-based face recognition with w-space identity-driven sampling.
\newblock {\em {IEEE} Trans. Biom. Behav. Identity Sci.}, 6(3):290--303, 2024.

\bibitem{DBLP:conf/icb/BoutrosKFKD23}
Fadi Boutros, Marcel Klemt, Meiling Fang, Arjan Kuijper, and Naser Damer.
\newblock Exfacegan: Exploring identity directions in gan's learned latent space for synthetic identity generation.
\newblock In {\em {IEEE} International Joint Conference on Biometrics, {IJCB} 2023, Ljubljana, Slovenia, September 25-28, 2023}, pages 1--10. {IEEE}, 2023.

\bibitem{DBLP:journals/ivc/BoutrosSFD23}
Fadi Boutros, Vitomir Struc, Julian Fi{\'{e}}rrez, and Naser Damer.
\newblock Synthetic data for face recognition: Current state and future prospects.
\newblock {\em Image Vis. Comput.}, 135:104688, 2023.

\bibitem{brown2020languagemodelsfewshotlearners}
Tom~B. Brown, Benjamin Mann, Nick Ryder, Melanie Subbiah, Jared Kaplan, Prafulla Dhariwal, Arvind Neelakantan, Pranav Shyam, Girish Sastry, Amanda Askell, Sandhini Agarwal, Ariel Herbert{-}Voss, Gretchen Krueger, Tom Henighan, Rewon Child, Aditya Ramesh, Daniel~M. Ziegler, Jeffrey Wu, Clemens Winter, Christopher Hesse, Mark Chen, Eric Sigler, Mateusz Litwin, Scott Gray, Benjamin Chess, Jack Clark, Christopher Berner, Sam McCandlish, Alec Radford, Ilya Sutskever, and Dario Amodei.
\newblock Language models are few-shot learners.
\newblock In {\em NeurIPS}, 2020.

\bibitem{DBLP:conf/fgr/CaoSXPZ18}
Qiong Cao, Li Shen, Weidi Xie, Omkar~M. Parkhi, and Andrew Zisserman.
\newblock Vggface2: {A} dataset for recognising faces across pose and age.
\newblock In {\em {FG}}, pages 67--74. {IEEE} Computer Society, 2018.

\bibitem{caron2021emergingpropertiesselfsupervisedvision}
Mathilde Caron, Hugo Touvron, Ishan Misra, Herv{\'{e}} J{\'{e}}gou, Julien Mairal, Piotr Bojanowski, and Armand Joulin.
\newblock Emerging properties in self-supervised vision transformers.
\newblock In {\em {ICCV}}, pages 9630--9640. {IEEE}, 2021.

\bibitem{Chen_2023_ICCV}
Feng Chen, Mario~Valerio Giuffrida, and Sotirios~A. Tsaftaris.
\newblock Adapting vision foundation models for plant phenotyping.
\newblock In {\em {ICCV} (Workshops)}, pages 604--613. {IEEE}, 2023.

\bibitem{AdaptFormer2022}
Shoufa Chen, Chongjian Ge, Zhan Tong, Jiangliu Wang, Yibing Song, Jue Wang, and Ping Luo.
\newblock Adaptformer: Adapting vision transformers for scalable visual recognition.
\newblock In {\em NeurIPS}, 2022.

\bibitem{chen2023visiontransformeradapterdense}
Zhe Chen, Yuchen Duan, Wenhai Wang, Junjun He, Tong Lu, Jifeng Dai, and Yu Qiao.
\newblock Vision transformer adapter for dense predictions.
\newblock In {\em {ICLR}}. OpenReview.net, 2023.

\bibitem{chowdhery2022palmscalinglanguagemodeling}
Aakanksha Chowdhery, Sharan Narang, Jacob Devlin, Maarten Bosma, Gaurav Mishra, Adam Roberts, Paul Barham, Hyung~Won Chung, Charles Sutton, Sebastian Gehrmann, Parker Schuh, Kensen Shi, Sasha Tsvyashchenko, Joshua Maynez, Abhishek Rao, Parker Barnes, Yi Tay, Noam Shazeer, Vinodkumar Prabhakaran, Emily Reif, Nan Du, Ben Hutchinson, Reiner Pope, James Bradbury, Jacob Austin, Michael Isard, Guy Gur{-}Ari, Pengcheng Yin, Toju Duke, Anselm Levskaya, Sanjay Ghemawat, Sunipa Dev, Henryk Michalewski, Xavier Garcia, Vedant Misra, Kevin Robinson, Liam Fedus, Denny Zhou, Daphne Ippolito, David Luan, Hyeontaek Lim, Barret Zoph, Alexander Spiridonov, Ryan Sepassi, David Dohan, Shivani Agrawal, Mark Omernick, Andrew~M. Dai, Thanumalayan~Sankaranarayana Pillai, Marie Pellat, Aitor Lewkowycz, Erica Moreira, Rewon Child, Oleksandr Polozov, Katherine Lee, Zongwei Zhou, Xuezhi Wang, Brennan Saeta, Mark Diaz, Orhan Firat, Michele Catasta, Jason Wei, Kathy Meier{-}Hellstern, Douglas Eck, Jeff Dean, Slav Petrov, and Noah Fiedel.
\newblock Palm: Scaling language modeling with pathways.
\newblock {\em J. Mach. Learn. Res.}, 24:240:1--240:113, 2023.

\bibitem{Randaugment_CVPR}
Ekin~D. Cubuk, Barret Zoph, Jonathon Shlens, and Quoc~V. Le.
\newblock Randaugment: Practical automated data augmentation with a reduced search space.
\newblock In {\em 2020 {IEEE/CVF} Conference on Computer Vision and Pattern Recognition, {CVPR} Workshops 2020, Seattle, WA, USA, June 14-19, 2020}, pages 3008--3017. Computer Vision Foundation / {IEEE}, 2020.

\bibitem{cui2024surgicaldinoadapterlearningfoundation}
Beilei Cui, Mobarakol Islam, Long Bai, and Hongliang Ren.
\newblock Surgical-dino: adapter learning of foundation models for depth estimation in endoscopic surgery.
\newblock {\em Int. J. Comput. Assist. Radiol. Surg.}, 19(6):1013--1020, 2024.

\bibitem{TransFace}
Jun Dan, Yang Liu, Haoyu Xie, Jiankang Deng, Haoran Xie, Xuansong Xie, and Baigui Sun.
\newblock Transface: Calibrating transformer training for face recognition from a data-centric perspective.
\newblock In {\em {IEEE/CVF} International Conference on Computer Vision, {ICCV} 2023, Paris, France, October 1-6, 2023}, pages 20585--20596. {IEEE}, 2023.

\bibitem{Deng_2022}
Jiankang Deng, Jia Guo, Jing Yang, Niannan Xue, Irene Kotsia, and Stefanos Zafeiriou.
\newblock Arcface: Additive angular margin loss for deep face recognition.
\newblock {\em IEEE Transactions on Pattern Analysis and Machine Intelligence}, 44(10):5962–5979, Oct. 2022.

\bibitem{DBLP:conf/iccvw/DengGZDLS19}
Jiankang Deng, Jia Guo, Debing Zhang, Yafeng Deng, Xiangju Lu, and Song Shi.
\newblock Lightweight face recognition challenge.
\newblock In {\em 2019 {IEEE/CVF} International Conference on Computer Vision Workshops, {ICCV} Workshops 2019, Seoul, Korea (South), October 27-28, 2019}, pages 2638--2646. {IEEE}, 2019.

\bibitem{devlin2019bertpretrainingdeepbidirectional}
Jacob Devlin, Ming{-}Wei Chang, Kenton Lee, and Kristina Toutanova.
\newblock {BERT:} pre-training of deep bidirectional transformers for language understanding.
\newblock In {\em {NAACL-HLT} {(1)}}, pages 4171--4186. Association for Computational Linguistics, 2019.

\bibitem{dosovitskiy2021imageworth16x16words}
Alexey Dosovitskiy, Lucas Beyer, Alexander Kolesnikov, Dirk Weissenborn, Xiaohua Zhai, Thomas Unterthiner, Mostafa Dehghani, Matthias Minderer, Georg Heigold, Sylvain Gelly, Jakob Uszkoreit, and Neil Houlsby.
\newblock An image is worth 16x16 words: Transformers for image recognition at scale.
\newblock In {\em {ICLR}}. OpenReview.net, 2021.

\bibitem{farmanifard2024irissamirissegmentationusing}
Parisa Farmanifard and Arun Ross.
\newblock Iris-sam: Iris segmentation using a foundational model.
\newblock {\em CoRR}, abs/2402.06497, 2024.

\bibitem{DBLP:conf/cvpr/Gong0021}
Sixue Gong, Xiaoming Liu, and Anil~K. Jain.
\newblock Mitigating face recognition bias via group adaptive classifier.
\newblock In {\em {IEEE} Conference on Computer Vision and Pattern Recognition, {CVPR} 2021, virtual, June 19-25, 2021}, pages 3414--3424. Computer Vision Foundation / {IEEE}, 2021.

\bibitem{gui2024surveyselfsupervisedlearningalgorithms}
Jie Gui, Tuo Chen, Jing Zhang, Qiong Cao, Zhenan Sun, Hao Luo, and Dacheng Tao.
\newblock A survey on self-supervised learning: Algorithms, applications, and future trends.
\newblock {\em {IEEE} Trans. Pattern Anal. Mach. Intell.}, 46(12):9052--9071, 2024.

\bibitem{guo2016ms}
Yandong Guo, Lei Zhang, Yuxiao Hu, Xiaodong He, and Jianfeng Gao.
\newblock Ms-celeb-1m: {A} dataset and benchmark for large-scale face recognition.
\newblock In Bastian Leibe, Jiri Matas, Nicu Sebe, and Max Welling, editors, {\em Computer Vision - {ECCV} 2016 - 14th European Conference, Amsterdam, The Netherlands, October 11-14, 2016, Proceedings, Part {III}}, volume 9907 of {\em Lecture Notes in Computer Science}, pages 87--102. Springer, 2016.

\bibitem{he2021maskedautoencodersscalablevision}
Kaiming He, Xinlei Chen, Saining Xie, Yanghao Li, Piotr Doll{\'{a}}r, and Ross~B. Girshick.
\newblock Masked autoencoders are scalable vision learners.
\newblock In {\em {CVPR}}, pages 15979--15988. {IEEE}, 2022.

\bibitem{He2015DeepRL}
Kaiming He, X. Zhang, Shaoqing Ren, and Jian Sun.
\newblock Deep residual learning for image recognition.
\newblock {\em 2016 IEEE Conference on Computer Vision and Pattern Recognition (CVPR)}, pages 770--778, 2015.

\bibitem{hu2021loralowrankadaptationlarge}
Edward~J. Hu, Yelong Shen, Phillip Wallis, Zeyuan Allen{-}Zhu, Yuanzhi Li, Shean Wang, Lu Wang, and Weizhu Chen.
\newblock Lora: Low-rank adaptation of large language models.
\newblock In {\em {ICLR}}. OpenReview.net, 2022.

\bibitem{huang:inria-00321923}
Gary~B. Huang, Marwan Mattar, Tamara Berg, and Eric Learned-Miller.
\newblock {Labeled Faces in the Wild: A Database forStudying Face Recognition in Unconstrained Environments}.
\newblock In {\em {Workshop on Faces in 'Real-Life' Images: Detection, Alignment, and Recognition}}, Marseille, France, Oct. 2008. {Erik Learned-Miller and Andras Ferencz and Fr{\'e}d{\'e}ric Jurie}.

\bibitem{huang2020curricularfaceadaptivecurriculumlearning}
Yuge Huang, Yuhan Wang, Ying Tai, Xiaoming Liu, Pengcheng Shen, Shaoxin Li, Jilin Li, and Feiyue Huang.
\newblock Curricularface: Adaptive curriculum learning loss for deep face recognition.
\newblock In {\em {CVPR}}, pages 5900--5909. Computer Vision Foundation / {IEEE}, 2020.

\bibitem{kalajdzievski2023rankstabilizationscalingfactor}
Damjan Kalajdzievski.
\newblock A rank stabilization scaling factor for fine-tuning with lora.
\newblock {\em CoRR}, abs/2312.03732, 2023.

\bibitem{kim2023adafacequalityadaptivemargin}
Minchul Kim, Anil~K. Jain, and Xiaoming Liu.
\newblock Adaface: Quality adaptive margin for face recognition.
\newblock In {\em {CVPR}}, pages 18729--18738. {IEEE}, 2022.

\bibitem{kim2023dcfacesyntheticfacegeneration}
Minchul Kim, Feng Liu, Anil~K. Jain, and Xiaoming Liu.
\newblock Dcface: Synthetic face generation with dual condition diffusion model.
\newblock In {\em {CVPR}}, pages 12715--12725. {IEEE}, 2023.

\bibitem{DBLP:conf/cvpr/KimS0JL24}
Minchul Kim, Yiyang Su, Feng Liu, Anil Jain, and Xiaoming Liu.
\newblock Keypoint relative position encoding for face recognition.
\newblock In {\em {CVPR}}, pages 244--255. {IEEE}, 2024.

\bibitem{kingma2017adammethodstochasticoptimization}
Diederik~P. Kingma and Jimmy Ba.
\newblock Adam: {A} method for stochastic optimization.
\newblock In {\em {ICLR} (Poster)}, 2015.

\bibitem{kirillov2023segment}
Alexander Kirillov, Eric Mintun, Nikhila Ravi, Hanzi Mao, Chlo{\'{e}} Rolland, Laura Gustafson, Tete Xiao, Spencer Whitehead, Alexander~C. Berg, Wan{-}Yen Lo, Piotr Doll{\'{a}}r, and Ross~B. Girshick.
\newblock Segment anything.
\newblock In {\em {ICCV}}, pages 3992--4003. {IEEE}, 2023.

\bibitem{9001031}
K.~S. Krishnapriya, Vítor Albiero, Kushal Vangara, Michael~C. King, and Kevin~W. Bowyer.
\newblock Issues related to face recognition accuracy varying based on race and skin tone.
\newblock {\em IEEE Transactions on Technology and Society}, 1(1):8--20, 2020.

\bibitem{DBLP:conf/icml/LiuWYY16}
Weiyang Liu, Yandong Wen, Zhiding Yu, and Meng Yang.
\newblock Large-margin softmax loss for convolutional neural networks.
\newblock In {\em {ICML}}, volume~48 of {\em {JMLR} Workshop and Conference Proceedings}, pages 507--516. JMLR.org, 2016.

\bibitem{loshchilov2017sgdrstochasticgradientdescent}
Ilya Loshchilov and Frank Hutter.
\newblock {SGDR:} stochastic gradient descent with warm restarts.
\newblock In {\em {ICLR} (Poster)}. OpenReview.net, 2017.

\bibitem{loshchilov2019decoupledweightdecayregularization}
Ilya Loshchilov and Frank Hutter.
\newblock Decoupled weight decay regularization.
\newblock In {\em {ICLR} (Poster)}. OpenReview.net, 2019.

\bibitem{DBLP:conf/icb/MazeADKMO0NACG18}
Brianna Maze, Jocelyn~C. Adams, James~A. Duncan, Nathan~D. Kalka, Tim Miller, Charles Otto, Anil~K. Jain, W.~Tyler Niggel, Janet Anderson, Jordan Cheney, and Patrick Grother.
\newblock {IARPA} janus benchmark - {C:} face dataset and protocol.
\newblock In {\em {ICB}}, pages 158--165. {IEEE}, 2018.

\bibitem{DBLP:journals/inffus/MelziTVKRLDMFOZZYZWLTKZDBVGFFMUG24}
Pietro Melzi, Ruben Tolosana, Rub{\'{e}}n Vera{-}Rodr{\'{\i}}guez, Minchul Kim, Christian Rathgeb, Xiaoming Liu, Ivan DeAndres{-}Tame, Aythami Morales, Julian Fi{\'{e}}rrez, Javier Ortega{-}Garcia, Weisong Zhao, Xiangyu Zhu, Zheyu Yan, Xiaoyu Zhang, Jinlin Wu, Zhen Lei, Suvidha Tripathi, Mahak Kothari, Md~Haider Zama, Debayan Deb, Bernardo Biesseck, Pedro Vidal, Roger Granada, Guilherme~P. Fickel, Gustavo F{\"{u}}hr, David Menotti, Alexander Unnervik, Anjith George, Christophe Ecabert, Hatef Otroshi{-}Shahreza, Parsa Rahimi, S{\'{e}}bastien Marcel, Ioannis Sarridis, Christos Koutlis, Georgia Baltsou, Symeon Papadopoulos, Christos Diou, Nicol{\`{o}}~Di Domenico, Guido Borghi, Lorenzo Pellegrini, Enrique Mas{-}Candela, {\'{A}}ngela S{\'{a}}nchez{-}P{\'{e}}rez, Andrea Atzori, Fadi Boutros, Naser Damer, Gianni Fenu, and Mirko Marras.
\newblock Frcsyn-ongoing: Benchmarking and comprehensive evaluation of real and synthetic data to improve face recognition systems.
\newblock {\em Inf. Fusion}, 107:102322, 2024.

\bibitem{meng2021magfaceuniversalrepresentationface}
Qiang Meng, Shichao Zhao, Zhida Huang, and Feng Zhou.
\newblock Magface: {A} universal representation for face recognition and quality assessment.
\newblock In {\em {CVPR}}, pages 14225--14234. Computer Vision Foundation / {IEEE}, 2021.

\bibitem{moschoglou2017agedb}
Stylianos Moschoglou, Athanasios Papaioannou, Christos Sagonas, Jiankang Deng, Irene Kotsia, and Stefanos Zafeiriou.
\newblock Agedb: the first manually collected, in-the-wild age database.
\newblock In {\em Proceedings of the IEEE Conference on Computer Vision and Pattern Recognition Workshop}, volume~2, page~5, 2017.

\bibitem{oquab2024DINOv2learningrobustvisual}
Maxime Oquab, Timoth{\'{e}}e Darcet, Th{\'{e}}o Moutakanni, Huy~V. Vo, Marc Szafraniec, Vasil Khalidov, Pierre Fernandez, Daniel Haziza, Francisco Massa, Alaaeldin El{-}Nouby, Mido Assran, Nicolas Ballas, Wojciech Galuba, Russell Howes, Po{-}Yao Huang, Shang{-}Wen Li, Ishan Misra, Michael Rabbat, Vasu Sharma, Gabriel Synnaeve, Hu Xu, Herv{\'{e}} J{\'{e}}gou, Julien Mairal, Patrick Labatut, Armand Joulin, and Piotr Bojanowski.
\newblock Dinov2: Learning robust visual features without supervision.
\newblock {\em Trans. Mach. Learn. Res.}, 2024, 2024.

\bibitem{DBLP:conf/fgr/Otroshi-Shahreza24}
Hatef Otroshi{-}Shahreza, Christophe Ecabert, Anjith George, Alexander Unnervik, S{\'{e}}bastien Marcel, Nicol{\`{o}}~Di Domenico, Guido Borghi, Davide Maltoni, Fadi Boutros, Julia Vogel, Naser Damer, {\'{A}}ngela S{\'{a}}nchez{-}P{\'{e}}rez, Enrique Mas{-}Candela, Jorge Calvo{-}Zaragoza, Bernardo Biesseck, Pedro Vidal, Roger Granada, David Menotti, Ivan DeAndres{-}Tame, Simone Maurizio~La Cava, Sara Concas, Pietro Melzi, Ruben Tolosana, Rub{\'{e}}n Vera{-}Rodr{\'{\i}}guez, Gianpaolo Perelli, Giulia Orr{\`{u}}, Gian~Luca Marcialis, and Julian Fi{\'{e}}rrez.
\newblock {SDFR:} synthetic data for face recognition competition.
\newblock In {\em 18th {IEEE} International Conference on Automatic Face and Gesture Recognition, {FG} 2024, Istanbul, Turkey, May 27-31, 2024}, pages 1--9. {IEEE}, 2024.

\bibitem{ozturk2024accuracybiasfacialhairstyle}
Kagan {\"{O}}zt{\"{u}}rk, Haiyu Wu, and Kevin~W. Bowyer.
\newblock Can the accuracy bias by facial hairstyle be reduced through balancing the training data?
\newblock In {\em {CVPR} Workshops}, pages 1519--1528. {IEEE}, 2024.

\bibitem{papantoniou2024arc2facefoundationmodelidconsistent}
Foivos~Paraperas Papantoniou, Alexandros Lattas, Stylianos Moschoglou, Jiankang Deng, Bernhard Kainz, and Stefanos Zafeiriou.
\newblock Arc2face: {A} foundation model for id-consistent human faces.
\newblock In {\em {ECCV} {(37)}}, volume 15095 of {\em Lecture Notes in Computer Science}, pages 241--261. Springer, 2024.

\bibitem{radford2021learningtransferablevisualmodels}
Alec Radford, Jong~Wook Kim, Chris Hallacy, Aditya Ramesh, Gabriel Goh, Sandhini Agarwal, Girish Sastry, Amanda Askell, Pamela Mishkin, Jack Clark, Gretchen Krueger, and Ilya Sutskever.
\newblock Learning transferable visual models from natural language supervision.
\newblock In {\em {ICML}}, volume 139 of {\em Proceedings of Machine Learning Research}, pages 8748--8763. {PMLR}, 2021.

\bibitem{ravi2024sam2segmentimages}
Nikhila Ravi, Valentin Gabeur, Yuan{-}Ting Hu, Ronghang Hu, Chaitanya Ryali, Tengyu Ma, Haitham Khedr, Roman R{\"{a}}dle, Chlo{\'{e}} Rolland, Laura Gustafson, Eric Mintun, Junting Pan, Kalyan~Vasudev Alwala, Nicolas Carion, Chao{-}Yuan Wu, Ross~B. Girshick, Piotr Doll{\'{a}}r, and Christoph Feichtenhofer.
\newblock {SAM} 2: Segment anything in images and videos.
\newblock {\em CoRR}, abs/2408.00714, 2024.

\bibitem{DBLP:conf/cvpr/SchroffKP15}
Florian Schroff, Dmitry Kalenichenko, and James Philbin.
\newblock Facenet: {A} unified embedding for face recognition and clustering.
\newblock In {\em {CVPR}}, pages 815--823. {IEEE} Computer Society, 2015.

\bibitem{c3517bca662f4193a58fd8f9e3145c8f}
Soumyadip Sengupta, {Jun Cheng} Chen, Carlos Castillo, {Vishal M.} Patel, Rama Chellappa, and {David W.} Jacobs.
\newblock Frontal to profile face verification in the wild.
\newblock In {\em 2016 IEEE Winter Conference on Applications of Computer Vision, WACV 2016}, 2016 IEEE Winter Conference on Applications of Computer Vision, WACV 2016. Institute of Electrical and Electronics Engineers Inc., May 2016.
\newblock Publisher Copyright: {\textcopyright} 2016 IEEE.; IEEE Winter Conference on Applications of Computer Vision, WACV 2016 ; Conference date: 07-03-2016 Through 10-03-2016.

\bibitem{DBLP:conf/nips/Sohn16}
Kihyuk Sohn.
\newblock Improved deep metric learning with multi-class n-pair loss objective.
\newblock In Daniel~D. Lee, Masashi Sugiyama, Ulrike von Luxburg, Isabelle Guyon, and Roman Garnett, editors, {\em Advances in Neural Information Processing Systems 29: Annual Conference on Neural Information Processing Systems 2016, December 5-10, 2016, Barcelona, Spain}, pages 1849--1857, 2016.

\bibitem{DBLP:conf/bmvc/SunT22}
Zhonglin Sun and Georgios Tzimiropoulos.
\newblock Part-based face recognition with vision transformers.
\newblock In {\em {BMVC}}, page 611. {BMVA} Press, 2022.

\bibitem{touvron2023llamaopenefficientfoundation}
Hugo Touvron, Thibaut Lavril, Gautier Izacard, Xavier Martinet, Marie{-}Anne Lachaux, Timoth{\'{e}}e Lacroix, Baptiste Rozi{\`{e}}re, Naman Goyal, Eric Hambro, Faisal Azhar, Aur{\'{e}}lien Rodriguez, Armand Joulin, Edouard Grave, and Guillaume Lample.
\newblock Llama: Open and efficient foundation language models.
\newblock {\em CoRR}, abs/2302.13971, 2023.

\bibitem{DBLP:journals/corr/abs-1906-06423}
Hugo Touvron, Andrea Vedaldi, Matthijs Douze, and Herv{\'{e}} J{\'{e}}gou.
\newblock Fixing the train-test resolution discrepancy.
\newblock In {\em NeurIPS}, pages 8250--8260, 2019.

\bibitem{wang2023sammeetsroboticsurgery}
An Wang, Mobarakol Islam, Mengya Xu, Yang Zhang, and Hongliang Ren.
\newblock {SAM} meets robotic surgery: An empirical study on generalization, robustness and adaptation.
\newblock In {\em ISIC/Care-AI/MedAGI/DeCaF@MICCAI}, volume 14393 of {\em Lecture Notes in Computer Science}, pages 234--244. Springer, 2023.

\bibitem{wang2018cosfacelargemargincosine}
Hao Wang, Yitong Wang, Zheng Zhou, Xing Ji, Dihong Gong, Jingchao Zhou, Zhifeng Li, and Wei Liu.
\newblock Cosface: Large margin cosine loss for deep face recognition.
\newblock In {\em {CVPR}}, pages 5265--5274. Computer Vision Foundation / {IEEE} Computer Society, 2018.

\bibitem{DBLP:journals/corr/abs-1911-10692}
Mei Wang and Weihong Deng.
\newblock Mitigating bias in face recognition using skewness-aware reinforcement learning.
\newblock In {\em Proceedings of the IEEE/CVF conference on computer vision and pattern recognition}, pages 9322--9331, 2020.

\bibitem{RFW}
Mei Wang, Weihong Deng, Jiani Hu, Xunqiang Tao, and Yaohai Huang.
\newblock Racial faces in the wild: Reducing racial bias by information maximization adaptation network.
\newblock In {\em 2019 {IEEE/CVF} International Conference on Computer Vision, {ICCV} 2019, Seoul, Korea (South), October 27 - November 2, 2019}, pages 692--702. {IEEE}, 2019.

\bibitem{PAMI_BIAS}
Mei Wang, Yaobin Zhang, and Weihong Deng.
\newblock Meta balanced network for fair face recognition.
\newblock {\em {IEEE} Trans. Pattern Anal. Mach. Intell.}, 44(11):8433--8448, 2022.

\bibitem{inproceedingsijbb}
Cameron Whitelam, Emma Taborsky, Austin Blanton, Brianna Maze, Jocelyn~C. Adams, Tim Miller, Nathan~D. Kalka, Anil~K. Jain, James~A. Duncan, Kristen Allen, Jordan Cheney, and Patrick Grother.
\newblock {IARPA} janus benchmark-b face dataset.
\newblock In {\em {CVPR} Workshops}, pages 592--600. {IEEE} Computer Society, 2017.

\bibitem{li2024id3identitypreservingyetdiversifieddiffusionmodels}
Jianqing Xu, Shen Li, Jiaying Wu, Miao Xiong, Ailin Deng, Jiazhen Ji, Yuge Huang, Guodong Mu, Wenjie Feng, Shouhong Ding, and Bryan Hooi.
\newblock \${\textbackslash}text\{{ID}\}{\textasciicircum}3\$: Identity-preserving-yet-diversified diffusion models for synthetic face recognition.
\newblock In {\em The Thirty-eighth Annual Conference on Neural Information Processing Systems}, 2024.

\bibitem{DBLP:journals/corr/YiLLL14a}
Dong Yi, Zhen Lei, Shengcai Liao, and Stan~Z. Li.
\newblock Learning face representation from scratch.
\newblock {\em CoRR}, abs/1411.7923, 2014.

\bibitem{Yucer2023RacialBW}
Seyma Yucer, Furkan Tektas, Noura~Al Moubayed, and Toby~P. Breckon.
\newblock Racial bias within face recognition: {A} survey.
\newblock {\em CoRR}, abs/2305.00817, 2023.

\bibitem{zanella2024lowrankfewshotadaptationvisionlanguage}
Maxime Zanella and Ismail~Ben Ayed.
\newblock Low-rank few-shot adaptation of vision-language models.
\newblock In {\em {CVPR} Workshops}, pages 1593--1603. {IEEE}, 2024.

\bibitem{zhang2024learningadaptfoundationmodel}
Bowen Zhang, Ying Chen, Long Bai, Yan Zhao, Yuxiang Sun, Yixuan Yuan, Jianhua Zhang, and Hongliang Ren.
\newblock Learning to adapt foundation model dinov2 for capsule endoscopy diagnosis.
\newblock {\em CoRR}, abs/2406.10508, 2024.

\bibitem{MTCNN}
Kaipeng Zhang, Zhanpeng Zhang, Zhifeng Li, and Yu Qiao.
\newblock Joint face detection and alignment using multitask cascaded convolutional networks.
\newblock {\em {IEEE} Signal Process. Lett.}, 23(10):1499--1503, 2016.

\bibitem{CPLFWTech}
T. Zheng and W. Deng.
\newblock Cross-pose lfw: A database for studying cross-pose face recognition in unconstrained environments.
\newblock Technical Report 18-01, Beijing University of Posts and Telecommunications, February 2018.

\bibitem{DBLP:journals/corr/abs-1708-08197}
Tianyue Zheng, Weihong Deng, and Jiani Hu.
\newblock Cross-age {LFW:} {A} database for studying cross-age face recognition in unconstrained environments.
\newblock {\em CoRR}, abs/1708.08197, 2017.

\bibitem{zhou2022ibotimagebertpretraining}
Jinghao Zhou, Chen Wei, Huiyu Wang, Wei Shen, Cihang Xie, Alan~L. Yuille, and Tao Kong.
\newblock ibot: Image {BERT} pre-training with online tokenizer.
\newblock {\em CoRR}, abs/2111.07832, 2021.

\bibitem{zhu2023understandingViTtrainsbadly}
Haoran Zhu, Boyuan Chen, and Carter Yang.
\newblock Understanding why vit trains badly on small datasets: An intuitive perspective.
\newblock {\em CoRR}, abs/2302.03751, 2023.

\bibitem{zhu2021webface260mbenchmarkunveilingpower}
Zheng Zhu, Guan Huang, Jiankang Deng, Yun Ye, Junjie Huang, Xinze Chen, Jiagang Zhu, Tian Yang, Jiwen Lu, Dalong Du, and Jie Zhou.
\newblock Webface260m: {A} benchmark unveiling the power of million-scale deep face recognition.
\newblock In {\em {CVPR}}, pages 10492--10502. Computer Vision Foundation / {IEEE}, 2021.

\end{thebibliography}
}

\end{document}